\documentclass[preprint,12pt,3p]{elsarticle}
\usepackage{amsmath}
\usepackage{amssymb}
\usepackage{graphicx}
\usepackage{multirow}
\usepackage{caption,subcaption}
\usepackage{url}
\usepackage{float}
\usepackage{xcolor}
\usepackage{booktabs}
\usepackage{fixltx2e}
\usepackage{hyperref}
\usepackage{adjustbox}     
\usepackage{setspace}
\usepackage[ruled,algonl,vlined]{algorithm2e}
\usepackage{fixmath}
\usepackage[toc,page]{appendix}

\newsavebox\CBox
\def\textBF#1{\sbox\CBox{#1}\resizebox{\wd\CBox}{\ht\CBox}{\textbf{#1}}}

\newcommand\inv[1]{#1\raisebox{1.5ex}{$\scriptscriptstyle-\!1$}}
\journal{Pattern Recognition}
\begin{document}
	\begin{frontmatter}
	\title{Group Preserving Label Embedding for Multi-Label Classification}
	\author[venk,curaj]{Vikas Kumar\corref{cor1}}
	\ead{vikas007bca@gmail.com}
	\author[venk,curaj]{Arun K Pujari}
	\ead{akpujari@curaj.ac.in}
    \author[venk]{Vineet Padmanabhan}
	\ead{vineetcs@uohyd.ernet.in}
	\author[venk]{Venkateswara Rao Kagita}
	\ead{venkateswar.rao.kagita@gmail.com}
	\cortext[cor1]{Corresponding author:Tel.: +91-40-23014559; fax: +91-40-23134012;}
	
	\address[venk]{Artificial Intelligence Lab, School of Computer and Information Sciences, \\ 
		University of Hyderabad, Hyderbad-500046, AndhraPradesh, India}	
	\address[curaj]{Central University of Rajasthan, Rajasthan, India}
	
	\begin{abstract}
		Multi-label	learning is concerned with the classification of data with multiple class labels. This is in contrast to the traditional classification problem	where every data instance has a single label. Due to the exponential size of output space, exploiting intrinsic information in feature and label spaces has been the major thrust of research in recent years and use of parametrization and embedding have been the prime focus. Researchers have studied several aspects of embedding which include label embedding, input embedding, dimensionality reduction and feature selection. These approaches differ from one another in their capability to capture other intrinsic properties such as label correlation, local invariance etc. We assume here that the input data form groups and as a result, the label matrix exhibits a sparsity pattern and hence the labels corresponding to objects in the same group have similar sparsity. In this paper, we study the embedding of labels together with the group information with an objective to build an efficient multi-label classification. We assume the existence of a  low-dimensional space onto which the feature vectors and label vectors can be embedded. In order to achieve this, we address three sub-problems namely; (1) Identification of groups of labels; (2) Embedding of label vectors to a low rank-space so that the sparsity characteristic of individual groups remains invariant; and (3) Determining a linear mapping that embeds the feature vectors onto the same set of points, as in stage 2, in the low-dimensional space.
		We compare our method with seven well-known algorithms on twelve benchmark data sets. Our experimental analysis manifests the superiority of our proposed method over state-of-art algorithms for multi-label learning. 
	\end{abstract}
    
   \end{frontmatter}
	\section{Introduction}
	
	Multi-label learning is concerned with the classification of data with multiple class labels. The objective of multi-label classification is to build a classifier that can automatically tag an example with the most relevant subset of labels. This problem can be seen as a generalization of \textit{single label} classification where an instance is associated with a unique class label from a set of disjoint labels $L$. The majority of the methods for supervised machine learning proceeds from a formal setting in which data objects (instances) are represented in the form of feature vectors. Thus, an instance $x$ is represented as a $D$ dimensional real-valued feature vector $(x_1, \ldots, x_D) \in \mathbb{R}^D$. In multi-label classification, each training example $x_i, 1\le j \le N$ is associated with a label vector $y_i \in  \{-1, 1\}^L$. The $+1$ entry at the $j$th coordinate of vector $y_i$ indicates the presence of label $j$ in data point $x_i$.  Given the pair, a feature matrix $X \in \mathbb{R}^{N \times D}$ and a label matrix $Y \in \{-1,1\}^{N \times L}$, the task of multi-label classification is to learn a parameterization $h : \mathbb{R}^D \rightarrow \{-1, 1\}^L$ that maps each instance (or, a feature vector)  to a set of  labels (a label vector). Multi-label classification has applications in many areas, such as machine learning~\cite{read2009classifier,wang2014enhancing}, computer vision~\cite{cabral2011matrix,boutell2004learning,wang2018learning}, and data mining~\cite{tsoumakas2007random,schapire2000boostexter}. 

	Existing methods of multi-label classification can be broadly divided into two categories~\cite{sorower2010literature,zhang2014review} - methods based on problem transformation and methods based on algorithm adaptation. The first approach transforms the multi-label	classification problem into one or more single-label classification or regression problems so that existing single label classification algorithms can be applied. During the past decade, number of techniques are proposed in the literature such as Binary Relevance (BR)~\cite{boutell2004learning}, Calibrated Label Ranking~\cite{furnkranz2008multilabel}, Classifier Chains~\cite{read2009classifier} and Random $k$-labelsets~\cite{tsoumakas2007random}. The second approach which is based on algorithm adaptation technique, extends or adapts specific learning algorithms to deal with multi-label data directly. Representative algorithms include AdaBoost.MH and AdaBoost.MR~\cite{schapire2000boostexter} which are two simple extensions of AdaBoost, ML-DT~\cite{clareknowledge} adapting decision tree techniques,  lazy learning techniques such as ML-kNN~\cite{zhang2007ml} and BR-kNN~\cite{spyromitros2008empirical} to name a few. 
	
	Machine learning is nowadays applied to massive data sets of considerable size, including potentially unbounded streams of data. Under such conditions, the scalability of learning algorithms is of major concern, calling for an effective data management and the use of appropriate data structures for time- and space-efficient implementations. 
	In the case of multi-label classification one of the major issues that arises when we have extremely large feature space and label space is that of \textit{scalability}.	 Most of the conventional algorithms of multi-label classification fail in such situations. To cope with the challenge of exponential-sized output space, use of  parametrization and embedding have been the prime focus. There are two major strategies of embedding for problems with large size output space. The first approach is to transform feature vectors from original feature space to an embedded space, which captures the intrinsic structure of the features. In~\cite{huang2016learning}, a parametrized approach is suggested to transform data from original feature space to label-specific feature space with the assumption that each class label is associated with sparse label-specific features. A joint learning framework in which feature space embedding and multi-label classification are performed simultaneously with low-rank constraints on embedding is proposed in~\cite{yu2014large}. The second approach is to transform the label vectors to an embedded space, followed by the association between feature vectors and embedded label space for classification purpose. With proper decoding process that maps the projected data  back to the original label space, the task of multi-label prediction is achieved~\cite{bi2013efficient,furnkranz2008multilabel,hsu2009multi}. We review the methods that fall into this category in Section~\ref{related Work}. There are attempts to learn embedding of both feature and label space simultaneously onto the same space. 
	Embedding of both features and labels to a single low-dimensional  space is no way obvious and cannot be a routine extension of feature-only embedding or label-only embedding. It is necessary to retain the intrinsic relationship while mapping feature- and label-vectors to a single space. In~\cite{qian2010semi}, semi-supervised dimensionality reduction is attempted for such embedding.  But dimensionality reduction techniques, a kind of nonlinear embedding, maintain some specific invariance and are not guaranteed to retain intrinsic relationship with feature vectors. As a result, though dimensionality reduction approaches are algorithmically feasible, their role is not justified in the present context. To retain the relationships of  feature and label spaces, a simultaneous nonlinear embedding is proposed in~\cite{kimura2016simultaneous}. Moreover, all these approaches do not yield results beyond a particular level of accuracy for problems with large data and large number of labels. Recent years have witnessed extensive applications of data mining, where features or data items are inherently organized into groups. In the context of  multi-label classification, there are few proposals which model group information but no detailed study in this direction has so far been undertaken. To exploit label-correlations in the data locally, it is assumed in~\cite{huang2012multi} that the training data are  in  groups with instances in the same group sharing the same label correlations. In~\cite{sun2016multi}, highly correlated labels are grouped together using the information from label and instance spaces and for every group, sparse meta-label-specific features are learnt. 
		
	The present research starts with the assumption that there exists a low-dimensional space onto which the given set of feature vectors and label vectors can be embedded. Feature vectors can be embedded as  points and label vectors correspond to linear predictors, as decision hyperplanes, in this embedded space. There are similarities among labels belonging to the same group such that their low-dimensional representations share the same sparsity pattern. For example, the set of labels in \textit{corel5k} data set can be grouped into \textit{landscape-nature}, \textit{humans}, \textit{food} etc.~\cite{fakhari2013combination}. The features such as \textit{eye} and \textit{leg} are specific to the \textit{humans} whereas the feature like \textit{ridge} is specific to the group \textit{landscape-nature}. Given the label matrix $Y$, it is necessary to find points in a space of reduced dimension and to determine decision hyperplanes such that the classification thereof is compatible with $Y$. While doing so, it is desirable that the process must retain group information of labels.  We achieve this by a matrix factorization based framework in which the label matrix $Y$ is approximated using the product of two matrices $U \in \mathbb{R}^{N \times d}$ and $V \in \mathbb{R}^{d \times L}$.  In a sense the row of matrix $U$ can be viewed as point in a reduced dimension and the column of $V$ defines a set of decision hyperplanes. If there is any dependency properties in labels of $Y$ (column  of $Y$), this is retained in dependencies in decision hyperplanes (column of $V$) and not in embedded points.	We use the $\ell_{2,1}$ norm regularization on $V$ to exploit the shared sparsity pattern among the label groups. The second sub-objective is to learn a linear mapping that maps the feature vectors onto the same set of points which are obtained as a result of factorization of label matrix. We achieve this by a separate optimization problem. We make use of correlation coefficients to capture the similarity relation and $\ell_1$ norm for regularization. We use FISTA~\cite{beck2009fast} type of method to learn the label embedding and subsequently mapping from feature space to the embedded label space. Thus, we develop a novel multi-label classification method in the present work.	To the best of our knowledge, there has not been any earlier attempt in this direction. We feel that this approach will eventually provide a robust classification technique as demonstrated by our experimental results which looks very promising.

	The rest of the paper is organized as follows. Section~\ref{related Work} briefly reviews the earlier research on multi-label learning. We introduce our proposed method, termed as GroPLE in Section~\ref{proposedMethod}. Experimental analysis of proposed method is reported in Section~\ref{experimentalSection}. Finally, Section~\ref{conclusion} concludes and indicates several issues for future work.
	
	\section{Related Work}
	\label{related Work}
	
	
	Label embedding (LE) is a popular strategy for multi-label classification where the aim is to embed the labels in a low-dimensional latent space via linear or local non-linear embeddings. LE algorithms transforms the original label vectors to an embedded space, followed by the association between input and embedded label space for classification purposes. With a proper decoding process which maps the projected data  back to the original label space, the task of multi-label prediction is achieved. Formally, given feature matrix $X$ and label matrix $Y$, the \textit{d}-dimensional embedding of the label space is typically found through a linear transformation matrix $W\in \mathcal{R}^{L \times d}$. A mapping $h : X \rightarrow YW$ is then learnt from feature space to reduced label space. We briefly review the major approaches of LE.
	
	The approach of Hsu et al.~\cite{hsu2009multi} projects the label vectors to a random low-dimensional space, fits a regression model in this space, then projects these predictions back to the original label space. In~\cite{tai2012multilabel}, principal component analysis (PCA) is employed on the label covariance matrix to extract a low-dimensional latent space. In~\cite{balasubramanian2012landmark},  a sparsity-regularized least square reconstruction objective is used to select a small set of  labels that can predict the remaining labels. 
	Recently, Yu et al.~\cite{yu2014large} and Jing et al.~\cite{jing2015semi} proposed to use trace norm regularization to identify a low-dimensional representation of the original large label space. Mineiro et al.~\cite{karampatziakis2015scalable} use randomized dimensionality reduction to learn a low-dimensional embedding that explicitly captures correlations between the instance features and their labels. Some methods work with label or feature similarity matrices, and seek to preserve the local structure of the data in the low-dimensional latent space.  Prabhu et al.~\cite{prabhu2014fastxml} propose a method to train a classification tree by minimizing the Normalized Discounted Cumulative Gain. Rai et al.~\cite{rai2015large} assume that the label vectors are generated by sampling from a weighted combination of label topics, where the mixture coefficients are determined by the instance features. Based on the assumption that all the output labels can be recovered by a small subset, multi-label classification via column subset selection approach (CSSP) is  proposed in~\cite{bi2013efficient}. Given a matrix $Y$, CSSP seeks to find a column index set $C \subset \{1,\dots, L\} $ with cardinality $d~(d  \ll L)$ so that columns with indices in $C$ can approximately span $Y$. The subset of columns is selected using a randomized sampling procedure. The problem is formulated as follows.    
	\begin{equation}
		\label{cssp}
		\underset{C}{min}~\|Y - Y_{C}Y_{C}^{\dag}Y\|_F
	\end{equation}
	where $\|\cdot\|_{F}$ is Frobenius norm, $Y_C$ denotes the submatrix consisting of columns of $Y$ with indices in $C$ and $Y_{C}Y_{C}^{\dag}$ is the projection matrix onto the $d$-dimensional space spanned by columns of $Y_C$. Alternatively, there have been emerging interests in recent multi-label methods that take the correlation information as prior knowledge while modeling the embedding (encoding). These methods can be efficient when the mapped label space has significantly lower dimensionality than the original label space~\cite{hsu2009multi}.

	In recent years, matrix factorization (MF) based approach is frequently used to achieve the LE which aims at determining two matrices $U \in \mathbb{R}^{ N \times d}$ and $V \in \mathbb{R}^{ d \times L}$. The matrix $U$ can be viewed as the \emph{basis matrix}, while the matrix $V$ can be treated as the \emph{coefficient matrix} and a common formulation  is the following optimization problem. 	
	\begin{equation} \label{basic-fact-formulation}
	\underset{U, V}{min} \; \ell(Y, U, V) + \lambda R(U, V)\\
	\end{equation}
	where  $\ell(\cdot)$ is a loss function that measures how well $UV$ approximates $Y$, $R(\cdot)$ is a regularization function that promotes various desired properties in $U$ and $V$ (sparsity, group-sparsity, etc.)  and the constant $\lambda \ge 0$ is the regularization parameter which controls the extent of regularization. In~\cite{lin2014multi}, a MF based approach is used to learn the label encoding and decoding matrix simultaneously. The problem is formulated as follows.
	\begin{equation}
		\label{faie}
		\underset{U, V}{min}~\|Y - UV\|_{F}^{2} + \alpha\Psi(X, U)
	\end{equation}
	where $U \in \mathcal{R}^{ N \times d}$ is the code matrix, $V \in \mathcal{R}^{ d \times L}$ is the decoding matrix, $\Psi(X,U)$ is used to make $U$ feature-aware by considering correlations between $X$ and $U$ as side information and the constant $\alpha \ge 0$ is the trade-off parameter. In order to reduce the noisy information in the label space, the method proposed in~\cite{jian2016multi} decompose the original space to a low-dimensional space. One encouraging property of this low-dimensional space is that most of the structures in the original output label space can be explained and recovered. Instead of globally projecting onto a linear low-dimensional subspace, the method proposed in~\cite{bhatia2015sparse} learns embeddings which non-linearly capture label correlations by preserving the pairwise distances between only the closest (rather than all) label vectors.

	\section{GroPLE: The Proposed Method}   
	\label{proposedMethod}	
	In this section, a novel method of multi-label classification is proposed. The proposed method GroPLE has three major stages namely (1) Identification of groups of labels; (2) Embedding of label vectors to a low rank-space so that the sparsity characteristic of individual groups remain invariant; and (3) Determining a linear mapping that embeds the feature vectors onto the same set of points, as in stage 2, in the low-dimensional space. \\
		
	\noindent \textbf{Identification of groups of labels:} The label groups are not given explicitly and it is necessary to learn from the label matrix $Y$. One approach is to cluster the columns of $Y$. There are several clustering algorithms proposed in the literature such as \textit{k-means}~\cite{jain1999data,jain2010data}, \textit{hierarchical clustering}~\cite{johnson1967hierarchical} and \textit{spectral clustering}~\cite{zelnik2004self,ng2002spectral,von2007tutorial}. We adopt \textit{spectral clustering}. We do not claim that \textit{spectral clustering} is the best option. We first construct a graph $G = <\mathcal{V}, E>$ in the label space, where $\mathcal{V}$ denotes the vertex/label set, and $E$ is the edge set containing edges between each label pair. We adopt heat kernel weight with self-tuning technique (for parameter $\sigma$) as edge weight if two labels are connected $A_{i,j} = \exp(\frac{(-\|Y_i - Y_j\|^2)}{\sigma})$ where $Y_i$ and $Y_j$ are the $i$th and $j$th column of matrix $Y$~\cite{zelnik2004self}. Labels can be grouped into $K$ clusters by performing	\textit{k-means} with K largest eigenvectors as seeds  of the normalized affinity matrix $L=D^{-\frac{1}{2}}AD^{-\frac{1}{2}}$, where $D$ is a diagonal matrix with $D_{i,i} = \sum_{j}{}A_{i,j}$.\\
		
	\noindent \textbf{Label Space Embedding:} Given a label matrix $Y$, each column corresponds to a label and our assumption is that related labels form groups.	
	Let the columns of $Y \in \{-1,1\}^{N \times L}$ are divided into K groups as $Y = (Y^1, \dots, Y^K)$, where $Y^k \in  \{-1, 1\}^{N \times L_k}$ and $\sum_{k}^{K} L_k = L$. Matrix factorization based approach of label embedding aims to find latent factor matrices $U$ and $V$ to approximate $Y$. In the present case, where labels are divided into groups, we approximate $Y^k$ using $U$ and $V^k$. Ideally, there should be a subset of columns of $U$ associated with any group and hence, the corresponding  vector in $V^k$ of a label should have nonzero values only for the entries which correspond to the subset of columns of $U$ associated with the group. More concretely, we expect that for deciding any label group all the features are not important and each label in that group can be decided by linear combination of fewer group features. To achieve this,  we add a $\ell_{2, 1}$-norm regularization on $V^k$ that encourages row sparsity of $V^k$. Then the sub-objective to learn the embedding from original label space is given by	
	\begin{equation}
	\label{GroPLE}
	\underset{U, V^1, \dots, V^K}{min}~f(U, V^1, \dots, V^K) = \sum_{k=1}^{K}\|Y^{k}-UV^{k}\|^2_F + \lambda_1\|U\|_F^2 + \lambda_2\sum_{k=1}^{K}\|V^k\|_{2,1}
	\end{equation}
	where for a given matrix $A \in \mathbb{R}^{n \times m}$, $\|A\|_F^2 = \sum_{i=1}^{n}{\sum_{j=1}^{m}{A_{ij}^2}}$ and $\|A\|_{2,1} = \sum_{i=1}^{n}{\sqrt{\sum_{j=1}^{m}A_{ij}^2}}$. We can solve Eq.~(\ref{GroPLE}) by alternating minimization scheme that iteratively optimizes each of the factor matrices keeping the other fixed. The details of derivation can be found in~\ref{LSDRderivation}. 
	
 	We feel that our proposal of group sparsity preserving label embedding may pave the way to overcome the difficulty in handling the infrequently occurring labels more efficiently. Due to the presence of infrequently occurring (tail) labels, the low-rank assumption does not hold in label embedding based multi-label classification~\cite{bhatia2015sparse,xu2016robust}. The proposed method can tackle the problem at two different stages - while identifying the groups of labels and while embedding of label vector to a low-dimensional space. In the first stage, the label grouping can be implemented either by adopting a technique in which the prior estimate of number of groups is not required or by tuning the number of groups parameter based on label distribution. In this stage, the set of infrequent labels may form one or more separate clusters and hence these can be handled separately in the label embedding phase. 
	On the other hand, if such an infrequent label is part of a group having frequently occurring labels,
	it might be sharing some common characteristics with frequently occurring labels in the group, which is handled by the optimization problem in Eq.~(\ref{GroPLE}).\\
	
	\noindent \textbf{Feature Space Embedding:}
	The $U$ matrix computed above as a result of the learning process represents a set of points and it is desired that these points, in some sense, represent the training objects. We assume that there exists a linear embedding $Z \in \mathbb{R}^{D \times d}$ that maps the feature matrix $X$ to $U$. Thus, we justify our hypothesis that there exists a low-dimensional space where both $X$ and $Y$ are embedded and this embedding retains the intrinsic feature-label relation as well as the group information. In order to achieve the embedding of feature vectors, we try to capture the correlation that exists in the embedded space and formulate the objective function as follows. 
	\begin{equation}
	\label{featureEmbedding}
	\underset{Z}{min}~ \|XZ - U\|_{F}^{2} + \alpha \sum_{j=1}^{d}R_{ij}Z_{i}^TZ_{j}  + \beta\|Z\|_{1}
	\end{equation}
	where $Z_i$ is the $i$th column of matrix $Z$ and $R_{ij} = 1 - C_{ij}$, where $C_{ij}$ represent the correlation coefficient between $i$th and $j$th column of matrix $U$. We employ LLSF~\cite{huang2015learning} to learn the transformation matrix $Z$.\\
	
	\noindent \textbf{Complexity Analysis:} We analyze the computational complexity of the proposed method. The time complexity of GroPLE mainly comprises of three components: formation of label groups and the optimization of the problem given in Eq.~(\ref{GroPLE}) and~(\ref{featureEmbedding}). The formation of label groups has two parts: construction of neighbourhood graph and spectral decomposition of a graph Laplacian. This part takes $O(NL^2 + L^3)$. For each iteration in Algorithm~\ref{algo:LE} (see ~\ref{LSDRderivation}), updating $U$ requires $O(NLd + d^3 + Nd^2)$. For simplicity of representation, we are ignoring the number of groups $K$ and using the total number of labels $L$. Hence, the updation of $V$ takes  $O(NLd + d^3 + Nd^2 + Ld^2)$. Let $t_1$ be the maximum number of iterations required for gradient update, then overall computation required in LE process is $O(NL^2 + L^3) + O(t_1(NLd + d^3 + Nd^2)) + O(t_1(NLd + d^3 + Nd^2 + Ld^2))$, that is, $O(t_1(NLd + d^3 + Nd^2 + Ld^2))$. Similarly, the complexity of feature space embedding is $O(ND^2 + D^3) + O(2NDd + Dd^2)$, that is, $O(t_2(2NDd + Dd^2))$, where $t_2$ is the number of iterations. Hence the overall computation required by GroPLE is $O(t_1(NLd + d^3 + Nd^2 + Ld^2) + t_2(2NDd + Dd^2))$.	  
	
	\section{Experimental Analysis}
	\label{experimentalSection}		
	To validate the proposed GroPLE, we perform experiments on twelve commonly used multi-label benchmark data sets. The detailed characteristics of these data sets are summarized in Table \ref{datasetsCharacteristics}. All the data sets are publicly available and can be downloaded from \textit{meka}\footnote{\href{http://meka.sourceforge.net/\#datasets}{http://meka.sourceforge.net/\#datasets}}
	and \textit{mulan}\footnote{\href{http://mulan.sourceforge.net/datasets-mlc.html}{http://mulan.sourceforge.net/datasets-mlc.html}}.
	
	\begin{table}[h!]
		\renewcommand{\arraystretch}{1}
		\scriptsize
		\centering
		\captionsetup{font=scriptsize,justification=centering}
		\caption{Description of the Experimental Data Sets.}
		\begin{tabular}{llllll}
			\toprule
			Data set&\#instance&\#Feature&\#Label&Domain&LC \\
			\hline
	    	genbase&662&1185&27&biology&1.252\\
			medical&978&1449&45&text&1.245\\
			CAL500&502&68&174&music&26.044\\
			corel5k&5000&499&374&image&3.522\\
			rcv1 (subset 1)&6000&944&101&text&2.880\\
			rcv1 (subset 2)&6000&944&101&text&2.634\\
			rcv1 (subset 3)&6000&944&101&text&2.614\\
			bibtex&7395&1836&159&text&2.402\\
			corle16k001&13766&500&153&image&2.859\\
			delicious&16105&500&983&text(web)&19.020\\
			mediamill&43907&120&101&video&4.376\\
			bookmarks&87856&2150&208&text&2.028\\
			\bottomrule
		\end{tabular}
		\label{datasetsCharacteristics}
	\end{table}    
	\subsection{Evaluation Metrics}
	To measure the performance of different algorithms, we have employed four evaluation metrics popularly used in multi-label classification, i.e. \textit{accuracy, example based $f_1$ measure, macro $f_1$} and \textit{micro $f_1$}~\cite{zhang2014review,sorower2010literature}. Given a test data set $\mathcal{D} = \{x_i, y_i~|~1 \le i\le N\}$, where $y_i \in \{-1, 1\}^L$ is the ground truth labels associated with the $i$th test example, and let $\hat{y_i}$ be its predicted set of labels.	
	
	\noindent \textbf{Accuracy} for an instance evaluates the proportion of correctly predicted labels to the total number of active(actual and predicted) labels for that instance. The overall accuracy for a data set is the average across all instances.
	\begin{equation*}
	Accuracy = \frac{1}{N}\sum_{i=1}^{N}\frac{|y_{i} \wedge \hat{y_{i}}|}{|y_{i} \vee \hat{y_{i}}|}
	\end{equation*}

	\noindent \textbf{Example based $F_1$ Measure} is the harmonic mean of precision and recall for each example.
	\begin{equation*}
	F_1 = \frac{1}{N}\sum_{i=1}^{N}\frac{2p_ir_i}{p_i + r_i}
	\end{equation*}
	where $p_i$ and $r_i$ are precision and recall for the $i$th example.
	\vspace{4pt}
	
	\noindent \textbf{Macro $F_1$ }is the harmonic mean of precision and recall for each label.
	\begin{equation*}
	Macro F_1 = \frac{1}{L}\sum_{i=1}^{L}\frac{2p_ir_i}{p_i + r_i}
	\end{equation*}
	where $p_i$ and $r_i$ are precision and recall for the $i$th label.
	\vspace{4pt}
	
	\noindent \textbf{Micro $F_1$} treats every entry of the label vector as an individual instance regardless of label distinction.
	\begin{equation*}
	Micro~F_1 = \frac{2\sum_{i=1}^{L}TP_i}{2\sum_{i=1}^{L}TP_i + \sum_{i=1}^{L}FP_i + \sum_{i=1}^{L}FN_i}
	\end{equation*}
	where $TP_i$, $FP_i$ and $FN_i$ are true positive, false positive and false negative for $i$th label, respectively.
	
	\subsection{Baseline Methods}
	
	For performance comparison, we consider seven well-known state-of-the-art algorithms and these are the following.
	
	\begin{itemize}
		\item \textbf{BSVM}~\cite{boutell2004learning}: This is one of the representative algorithms of problem transformation methods, which treat each label as separate binary classification problem. For every label, an independent binary classifier is trained by considering the examples with given class label as positive and others as negative. LIBSVM~\cite{CC01a} is employed as the binary learner for classifier induction to instantiate BSVM.
		\item \textbf{LLSF}~\cite{tai2012multilabel}: This method addresses the inconsistency problem in multi-label classification by learning label specific features for the discrimination of each class label.  
		\item \textbf{PLST}~\cite{tai2012multilabel}: Principal label space transformation (PLST) uses singular value decomposition (SVD) to project the original label space into a low dimensional label space. 
		\item \textbf{CPLST}~\cite{chen2012feature}: CPLST is a feature-aware conditional principal label space transformation which utilizes the feature information during label embedding. 
		\item \textbf{FAiE}~\cite{lin2014multi}: FAiE encodes the original label space to a low-dimensional latent space via feature-aware implicit label space encoding. It directly learns a feature-aware code matrix and a linear decoding matrix via jointly maximizing recoverability of the original label space. 
		\item \textbf{LEML}~\cite{yu2014large}:  In this method a framework is developed to model  multi-label classification as  generic empirical risk minimization (ERM) problem with low-rank constraint on linear transformation. It can also be seen as a joint learning framework in which dimensionality reduction and multi-label classification are performed simultaneously. 
		\item \textbf{MLSF}~\cite{sun2016multi}: Based on the assumption that meta-labels with specific features exist in the scenario of multi-label classification, MLSF embed label correlations into meta-labels in such a way that the member labels in a meta-label share strong dependency with each other but have weak dependency with the other non-member labels. 
	\end{itemize}
	 A linear ridge regression model is used in PLST, CPLST and FAiE to learn the association between feature space and reduced label space. For PLST, CPLST, FAiE  and LEML the number of reduced dimensions $d$ is searched in  $\{\lceil0.1L\rceil, \lceil0.2L\rceil,\dots, \lceil0.8L\rceil\}$.   The regularization parameter in ridge regression, the parameter $\alpha$ in FAiE, the parameter $\lambda$ in LEML and the parameters $\alpha$ and $\beta$ in LLSF are searched in the range $\{ 10^{-4}, 10^{-3}, \dots, 10^{4}\}$. For MLSF, the number of meta-labels $K$ is searched in  $\{\lceil L/5 \rceil, \lceil L/10 \rceil, \lceil L/15 \rceil, \lceil L/20 \rceil \}$ and the parameters $\gamma$ and $\rho$ are tuned from the candidate set $\{ 10^{-4}, 10^{-3}, \dots, 10^{4}\}$. The remaining hyper-parameters  were kept fixed across all datasets as was done in~\cite{sun2016multi}.	Implementations of LLSF, PLST, CPLST, FAiE, LEML and MLSF were provided by the authors.
	
	\subsection{Results and Discussion}
\begin{figure}[ht!]
    	\centering
    	\includegraphics[width=3.3in,height=3in]{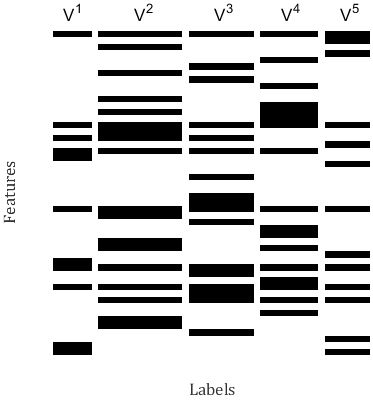}
    	\caption{Latent factor matrix $V^k$ recovered with five label groups.} 
    	\label{grayScaleImgofLatentFactor}
    \end{figure}

     We first demonstrate the effect of group sparsity regularization to support our hypothesis that there exists label groups and labels belonging to the same group share similar sparsity pattern (feature representation) in their latent factor representation. We have therefore performed an experiment on the \emph{Medical} data set. 
     As discussed previously,  in the real- world data sets, the label groups are not given explicitly and it is necessary to learn from the label matrix $Y$.  We have employed the method described in Section~\ref{proposedMethod} to divide the label matrix $Y$ into five groups. The feature matrix $V^k$, $1\le k \le 5$, for each group is learnt using the procedure given in Algorithm~\ref{algo:LE}. For this experiment, the regularization parameter $\lambda_1$ and $\lambda_2$ in Eq.~(\ref{GroPLE}) are selected using \textit{cross-validation}. We plotted the recovered feature matrix $V^k$ for each group in Figure~\ref{grayScaleImgofLatentFactor}. For simplicity of representation, the non-zero rows in the recovered matrix $V^k$ are shown as shaded rows and a gap is artificially created to distinguish between different $V^k$'s. It is evident from Figure~\ref{grayScaleImgofLatentFactor} that the labels corresponding to the same group have similar feature representation and in case if a feature is not present in a group, all the labels have the corresponding feature entries as zero.  It can also be seen that the feature matrix recovered for different groups exhibits a different sparsity pattern.     
                          
    To study the sensitivity of GroPLE with respect to the regularization parameters $\lambda_1$ and $\lambda_2$, we have conducted experiments  on \textit{Medical} and \textit{Genbase} data sets. We perform \textit{five-fold cross validation} on each data set and the mean value of accuracy is recorded.   In this experiment, the latent dimension space $d$ is fixed to $100$, the number of groups $K$ is fixed to $5$ and the regularization parameters $\lambda_1$ and $\lambda_2$ are searched in  $\{10^{-3}, 10^{-2}, \dots, 10^{2}\}$.  
    For each $(\lambda_1, \lambda_2)$ -pair, the regularization parameters $\alpha$, $\beta$ are searched in $\{ 10^{-4}, 10^{-3}, \dots, 10^{4}\}$.  
     \begin{figure}[ht!]
    	\centering
    	\begin{subfigure}{0.5\textwidth}
    		\centering
    		\includegraphics[width=3.3in,height=3in]{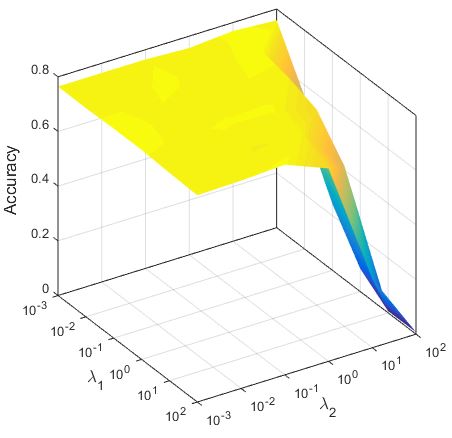}
    		\caption{Medical dataset}
    		\label{fig:lambda1lambda2Accuracy_Medical}
    	\end{subfigure}%
    	\begin{subfigure}{0.5\textwidth}
    		\centering
    		\includegraphics[width=3.3in,height=3in]{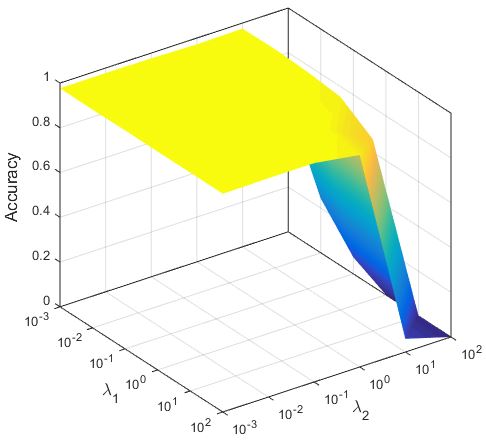}
    		\caption{Genbase dataset}
    		\label{fig:lambda1lambda2Accuracy_Genbase}
    	\end{subfigure}
    	\caption{Influence of regularization parameters $\lambda_1$ and $\lambda_2$.}		
    	\label{lambda1Lambda2Analysis}
    \end{figure}    
    Figure~\ref{fig:lambda1lambda2Accuracy_Medical} and~\ref{fig:lambda1lambda2Accuracy_Genbase} report the influence of parameters $\lambda_1$ and $\lambda_2$ on  \textit{Medical} and \textit{Genbase} data set, respectively.  It can be seen from Figure~\ref{lambda1Lambda2Analysis} that in most cases: (a) GroPLE perform worse when the value of $\lambda_1$ is large; (b) 
    The performance of GroPLE is stable with the different values of group sparsity regularization $\lambda_2$, but larger values such as $\lambda_2 > 1$ is often harmful.
    Therefore, we fixed the regularization parameter $\lambda_1$ and $\lambda_2$ to $0.001$ and $1$, respectively, for the subsequent experiments.
   
\begin{figure}[h!]
	\begin{subfigure}{0.5\textwidth}
		\centering
		\includegraphics[width=3.24in,height=3in]{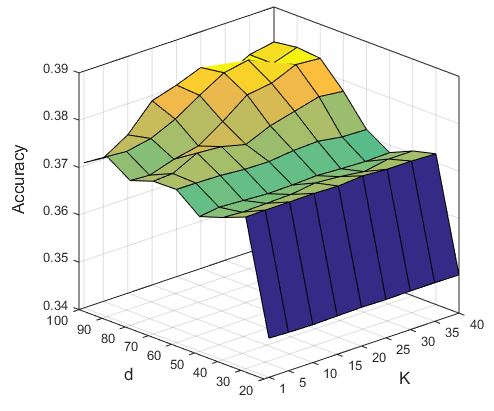}
		\caption{Accuracy}
		\label{fig:groupVsNoGroupAccuracy}
	\end{subfigure}%
	\begin{subfigure}{0.5\textwidth}
		\centering
		\includegraphics[width=3.24in,height=3in]{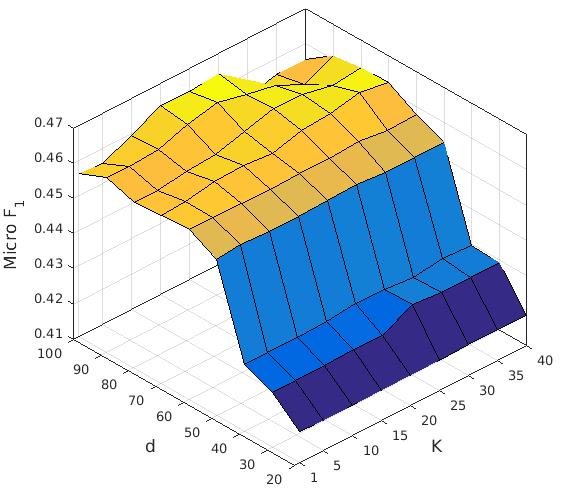}
		\caption{Micro $F_1$}
		\label{fig:groupVsNoGroupMicroF1}
	\end{subfigure}		
	\caption{Performance of GroPLE on \textit{rcv1 (subset 2)} data set with different group size.}
	\label{groupVsNoGroup}
\end{figure}
  We have also analyzed the performance of GroPLE with respect to the latent dimension $d$ and the number of groups $K$ on \textit{rcv1 (subset 2)} data set.  We have conducted \textit{five-fold cross validation} and the mean value of accuracy is recorded.  The latent dimension $d$ is varied from $[20,100]$ with step size $10$ and the number of groups $K$ is selected from $\{1, 5, 10, 15, 20, \dots, 40\}$.
  \begin{table*}[ht!]
  	\renewcommand{\arraystretch}{1.26}
  	\centering	
  	\captionsetup{font=scriptsize,justification=centering}
  	\caption{Experimental results of each comparing algorithm (mean$\pm$std rank)  in terms of Accuracy, Example Based $F_1$, Macro $F_1$, and Micro $F_1$. Method that cannot be run  with available resources are denoted as ``-".}
  	\adjustbox{max width=\linewidth}{
  		\begin{tabular}{lclclclclclclclclclcl}
  			\toprule
  			\multirow{2}{*}{Data set}&\multicolumn{15}{c}{Accuracy}\\ \cline{2-17}
  			&GroPLE&&LLSF&&PLST&&CPLST&&FAiE&&LEML&&MLSF&&BSVM& \\
  			\hline
  			genbase&0.972 $\pm$ 0.014&3&0.971 $\pm$ 0.015&5&0.968 $\pm$ 0.013&7&0.971 $\pm$ 0.013&5&0.965 $\pm$ 0.014&8&0.971 $\pm$ 0.014&5&0.973 $\pm$ 0.015&2&\textBF{0.976 $\pm$ 0.010}&\textBF{1}\\
  			medical&0.765 $\pm$ 0.026&2.5&0.762 $\pm$ 0.035&5.5&0.762 $\pm$ 0.034&5.5&0.763 $\pm$ 0.033&4&0.765 $\pm$ 0.034&2.5&0.690 $\pm$ 0.025&8&\textBF{0.773 $\pm$ 0.026}&\textBF{1}&0.753 $\pm$ 0.034&7\\
  			CAL500&0.233 $\pm$ 0.010&2&0.222 $\pm$ 0.009&5.5&0.224 $\pm$ 0.007&3.5&0.224 $\pm$ 0.007&3.5&\textBF{0.235 $\pm$ 0.007}&\textBF{1}&0.222 $\pm$ 0.006&5.5&0.165 $\pm$ 0.036&8&0.202 $\pm$ 0.007&7\\
  			corel5k&\textBF{0.150 $\pm$ 0.002}&\textBF{1}&0.120 $\pm$ 0.002&2&0.054 $\pm$ 0.002&6.5&0.054 $\pm$ 0.001&6.5&0.089 $\pm$ 0.003&3&0.053 $\pm$ 0.002&8&0.086 $\pm$ 0.007&4& 0.081 $\pm$ 0.002&5\\
  			rcv1 (subset 1)&\textBF{0.333 $\pm$ 0.003}&\textBF{1}&0.297 $\pm$ 0.009&3&0.231 $\pm$ 0.009&7&0.232 $\pm$ 0.008&6&0.265 $\pm$ 0.008&5&0.223 $\pm$ 0.006&8&0.320 $\pm$ 0.010&2&0.288 $\pm$ 0.010&4\\
  			rcv1 (subset 2)&\textBF{0.374 $\pm$ 0.008}&\textBF{1}&0.340 $\pm$ 0.012&4&0.299 $\pm$ 0.014&6.5&0.299 $\pm$ 0.013&6.5&0.328 $\pm$ 0.011&5&0.286 $\pm$ 0.013&8&0.371 $\pm$ 0.012&2&0.348 $\pm$ 0.012&3\\
  			rcv1 (subset 3)&0.376 $\pm$ 0.015&2&0.296 $\pm$ 0.008&5&0.288 $\pm$ 0.014&6.5&0.288 $\pm$ 0.015&6.5&0.319 $\pm$ 0.018&4&0.280 $\pm$ 0.014&8&\textBF{0.380 $\pm$ 0.013}&\textBF{1}&0.347 $\pm$ 0.009&3\\
  			bibtex&0.330 $\pm$ 0.005&2&\textBF{0.350 $\pm$ 0.005}&\textBF{1}&0.284 $\pm$ 0.009&8&0.288 $\pm$ 0.007&6&0.310 $\pm$ 0.004&5&0.287 $\pm$ 0.007&7&0.328 $\pm$ 0.008&3.5&0.328 $\pm$ 0.008&3.5\\
  			corel16k001&0.158 $\pm$ 0.006&2&\textBF{0.166 $\pm$ 0.005}&\textBF{1}&0.038 $\pm$ 0.001&5.5&0.038 $\pm$ 0.001&5.5&0.080 $\pm$ 0.002&3&0.039 $\pm$ 0.002&4&0.025 $\pm$ 0.004&7.5&0.025 $\pm$ 0.002&7.5\\
  			delicious&\textBF{0.185 $\pm$ 0.001}&\textBF{1}&0.153 $\pm$ 0.001&2&0.109 $\pm$ 0.001&6.5&0.109 $\pm$ 0.002&6.5&0.134 $\pm$ 0.001&3&0.093 $\pm$ 0.001&8&0.120 $\pm$ 0.008&5&0.130 $\pm$ 0.001&4\\
  			mediamill&\textBF{0.434 $\pm$ 0.004}&\textBF{1}&0.412 $\pm$ 0.003&5&0.414 $\pm$ 0.003&3.5&0.414 $\pm$ 0.003&3.5&0.425 $\pm$ 0.003&2&0.411 $\pm$ 0.003&6&0.379 $\pm$ 0.017&8&0.393 $\pm$ 0.004&7\\
  			bookmarks&\textBF{0.281 $\pm$ 0.004}&&0.246 $\pm$ 0.002&&0.174 $\pm$ 0.002&&\textBF{$-$}&&\textBF{$-$}&&0.174 $\pm$ 0.002&&0.163 $\pm$ 0.004&&\textBF{$-$}&\\
  			\hline
  			\multirow{2}{*}{Data set}&\multicolumn{15}{c}{Example based $F_1$}\\ \cline{2-17}
  			&GroPLE&&LLSF&&PLST&&CPLST&&FAiE&&LEML&&MLSF&&BSVM& \\
  			\hline
  			genbase&0.978 $\pm$ 0.013&2&0.977 $\pm$ 0.014&4.5&0.975 $\pm$ 0.012&7&0.977 $\pm$ 0.013&4.5&0.973 $\pm$ 0.012&8&0.977 $\pm$ 0.013&4.5&0.977 $\pm$ 0.014&4.5&\textBF{0.980 $\pm$ 0.010}&\textBF{1}\\
  			medical&0.796 $\pm$ 0.025&2.5&0.795 $\pm$ 0.034&4&0.793 $\pm$ 0.034&6&0.794 $\pm$ 0.033&5&0.796 $\pm$ 0.034&2.5&0.738 $\pm$ 0.027&8&\textBF{0.799 $\pm$ 0.023}&\textBF{1}&0.783 $\pm$ 0.032&7\\
  			CAL500&\textBF{0.369 $\pm$ 0.013}&\textBF{1.5}&0.354 $\pm$ 0.011&6&0.357 $\pm$ 0.009&3.5&0.357 $\pm$ 0.009&3&\textBF{0.369 $\pm$ 0.009}&\textBF{1.5}&0.355 $\pm$ 0.008&5&0.277 $\pm$ 0.053&8&0.330 $\pm$ 0.010&7\\
  			corel5k&\textBF{0.229 $\pm$ 0.002}&\textBF{1}&0.168 $\pm$ 0.002&2&0.078 $\pm$ 0.002&6.5&0.078 $\pm$ 0.002&6
  			.5&0.127 $\pm$ 0.004&3&0.076 $\pm$ 0.002&8&0.119 $\pm$ 0.009&4&0.114 $\pm$ 0.004&5\\
  			rcv1 (subset 1)&\textBF{0.446 $\pm$ 0.004}&\textBF{1}&0.384 $\pm$ 0.010&3&0.298 $\pm$ 0.009 &7&0.299 $\pm$ 0.008&6&0.339 $\pm$ 0.009&5&0.288 $\pm$ 0.006&8&0.395 $\pm$ 0.009&2&0.379 $\pm$ 0.012&4\\
  			rcv1 (subset 2)&\textBF{0.468 $\pm$ 0.010}&\textBF{1}&0.398 $\pm$ 0.013&4&0.346 $\pm$ 0.015&6.5&0.346 $\pm$ 0.014&6.5&0.380 $\pm$ 0.011&5&0.331 $\pm$ 0.014&8&0.433 $\pm$ 0.011&2&0.423 $\pm$ 0.012&3\\
  			rcv1 (subset 3)&\textBF{0.471 $\pm$ 0.013}&\textBF{1}&0.397 $\pm$ 0.008&4& 0.335 $\pm$ 0.015&6&0.334 $\pm$ 0.015&7&0.371 $\pm$ 0.018&5&0.325 $\pm$ 0.014&8&0.440 $\pm$ 0.013&2&0.422 $\pm$ 0.008&3\\
  			bibtex&0.415 $\pm$ 0.003&2&\textBF{0.426 $\pm$ 0.005}&\textBF{1}&0.335 $\pm$ 0.009&8&0.339 $\pm$ 0.008&6&0.367 $\pm$ 0.004&5&0.338 $\pm$ 0.008&7&0.397 $\pm$ 0.008&4&0.400 $\pm$ 0.008&3\\
  			corel16k001&0.227 $\pm$ 0.008&2&\textBF{0.243 $\pm$ 0.007}&\textBF{1}&0.053 $\pm$ 0.002&5&0.053 $\pm$ 0.002&5&0.111 $\pm$ 0.002&3&0.053 $\pm$ 0.003&5&0.033 $\pm$ 0.004&8&0.034 $\pm$ 0.002&7\\
  			delicious&\textBF{0.293 $\pm$ 0.001}&\textBF{1}&0.237 $\pm$ 0.002&2&0.169 $\pm$ 0.002&6.5&0.169 $\pm$ 0.002&6.5&0.209 $\pm$ 0.001&3&0.142 $\pm$ 0.002&8&0.184 $\pm$ 0.012&5&0.201 $\pm$ 0.002&4\\
  			mediamill&\textBF{0.553 $\pm$ 0.004}&\textBF{1}&0.532 $\pm$ 0.004&5&0.533 $\pm$ 0.004&3.5&0.533 $\pm$ 0.004&3.5&0.544 $\pm$ 0.004&2&0.530 $\pm$ 0.003&6&0.489 $\pm$ 0.021&8&0.515 $\pm$ 0.004&7\\
  			bookmarks&\textBF{0.322 $\pm$ 0.006}&&0.271 $\pm$ 0.001&&\textBF{$-$}&&0.179 $\pm$ 0.002&&\textBF{$-$}&&0.180 $\pm$ 0.002&&0.166 $\pm$ 0.003&&\textBF{$-$}&\\
  			\hline
  			\multirow{2}{*}{Data set}&\multicolumn{15}{c}{Macro $F_1$}\\ \cline{2-17}
  			&GroPLE&&LLSF&&PLST&&CPLST&&FAiE&&LEML&&MLSF&&BSVM& \\
  			\hline
  			genbase&0.710 $\pm$ 0.087&5&0.731 $\pm$ 0.088&2&0.709 $\pm$ 0.068&6&0.711 $\pm$ 0.075&4&0.682 $\pm$ 0.066&8&0.704 $\pm$ 0.077&7&0.728 $\pm$ 0.091&3&\textBF{0.737 $\pm$ 0.082}&1\\
  			medical&0.369 $\pm$ 0.026&6&0.354 $\pm$ 0.034&7&0.373 $\pm$ 0.025&5&0.378 $\pm$ 0.030&4&0.383 $\pm$ 0.025&2.5&0.342 $\pm$ 0.025&8&\textBF{0.402 $\pm$ 0.044}&\textBF{1}&0.383 $\pm$ 0.037&2.5\\
  			CAL500&\textBF{0.133 $\pm$ 0.008}&\textBF{1}&0.102 $\pm$ 0.006&6&0.110 $\pm$ 0.006&3&0.104 $\pm$ 0.005&5&0.118 $\pm$ 0.004&2&0.109 $\pm$ 0.005&4&0.032 $\pm$ 0.008&8&0.057 $\pm$ 0.001&7\\
  			corel5k&\textBF{0.048 $\pm$ 0.002}&\textBF{1}&0.036 $\pm$ 0.002&4&0.016 $\pm$ 0.002&6.5&0.016 $\pm$ 0.001&6.5&0.026 $\pm$ 0.001&5&0.015 $\pm$ 0.002&8&0.046 $\pm$ 0.003&2&0.044 $\pm$ 0.002&3\\
  			rcv1 (subset 1)&\textBF{0.262 $\pm$ 0.007}&\textBF{1}&0.209 $\pm$ 0.009&4&0.126 $\pm$ 0.006&6&0.125 $\pm$ 0.005&7.5&0.163 $\pm$ 0.009&5&0.125 $\pm$ 0.005&7.5&0.255 $\pm$ 0.008&3&0.257 $\pm$ 0.005&2\\
  			rcv1 (subset 2)&\textBF{0.250 $\pm$ 0.008}&\textBF{1}&0.175 $\pm$ 0.009&4&0.111 $\pm$ 0.006&7.5&0.112 $\pm$ 0.006&6&0.148 $\pm$ 0.004&5&0.111 $\pm$ 0.006&7.5&0.241 $\pm$ 0.013&2&0.240 $\pm$ 0.007&3\\
  			rcv1 (subset 3)&\textBF{0.244 $\pm$ 0.003}&\textBF{1}&0.230 $\pm$ 0.011&4&0.109 $\pm$ 0.003&6&0.105 $\pm$ 0.004&8&0.143 $\pm$ 0.007&5&0.108 $\pm$ 0.004&7&0.239 $\pm$ 0.016&2&0.231 $\pm$ 0.012&3\\
  			bibtex&0.304 $\pm$ 0.014&4&\textBF{0.343 $\pm$ 0.008}&\textBF{1}& 0.197 $\pm$ 0.005&8&0.208 $\pm$ 0.008&6.5&0.236 $\pm$ 0.007&5&0.208 $\pm$ 0.008&6.5&0.323 $\pm$ 0.006&3&0.327 $\pm$ 0.004&2\\
  			corel16k001&\textBF{0.088 $\pm$ 0.004}&\textBF{1}&0.079 $\pm$ 0.006&2&0.015 $\pm$ 0.001&7&0.015 $\pm$ 0.002&7&0.023 $\pm$ 0.002&5&0.015 $\pm$ 0.002&7&0.040 $\pm$ 0.009&3&0.036 $\pm$ 0.005&4\\
  			delicious&0.089 $\pm$ 0.001&4&0.095 $\pm$ 0.001&3&0.048 $\pm$ 0.001&7&0.048 $\pm$ 0.002&7&0.059 $\pm$ 0.002&5&0.048 $\pm$ 0.002&7&\textBF{0.101 $\pm$ 0.002}&\textBF{1}&0.100 $\pm$ 0.003&2\\
  			mediamill&\textBF{0.086 $\pm$ 0.001}&\textBF{1}&0.044 $\pm$ 0.000&5&0.045 $\pm$ 0.000&3.5&0.045 $\pm$ 0.001&3.5&0.055 $\pm$ 0.001&2&0.043 $\pm$ 0.000&6&0.031 $\pm$ 0.008&8&0.032 $\pm$ 0.001&7\\
  			bookmarks&0.138 $\pm$ 0.007&&\textBF{0.158 $\pm$ 0.002}&&0.057 $\pm$ 0.001&&\textBF{$-$}&&\textBF{$-$}&&0.059 $\pm$ 0.001&&0.043 $\pm$ 0.002&&\textBF{$-$}&\\
  			\hline
  			\multirow{2}{*}{Data set}&\multicolumn{15}{c}{Micro $F_1$}\\ \cline{2-17}
  			&GroPLE&&LLSF&&PLST&&CPLST&&FAiE&&LEML&&MLSF&&BSVM& \\
  			\hline
  			genbase&0.957 $\pm$ 0.030&8&0.971 $\pm$ 0.014&5&0.969 $\pm$ 0.012&6&0.972 $\pm$ 0.013&4&0.967 $\pm$ 0.013&7&0.973 $\pm$ 0.013&3&0.977 $\pm$ 0.011&2&\textBF{0.979 $\pm$ 0.008}&1\\
  			medical&0.819 $\pm$ 0.025&4&0.813 $\pm$ 0.030&6&0.821 $\pm$ 0.030&3&0.822 $\pm$ 0.029&2&\textBF{0.823 $\pm$ 0.030}&\textBF{1}&0.739 $\pm$ 0.018&8&0.817 $\pm$ 0.020&5&0.812 $\pm$ 0.027&7\\
  			CAL500&\textBF{0.374 $\pm$ 0.012}&\textBF{1.5}&0.358 $\pm$ 0.011&6&0.360 $\pm$ 0.008&4&0.361 $\pm$ 0.009&3&\textBF{0.374 $\pm$ 0.008}&\textBF{1.5}&0.359 $\pm$ 0.007&5&0.271 $\pm$ 0.052&8&0.327 $\pm$ 0.009&7\\
  			corel5k&\textBF{0.241 $\pm$ 0.003}&\textBF{1}&0.215 $\pm$ 0.004&2&0.105 $\pm$ 0.003&6.5&0.105 $\pm$ 0.002&6.5&0.162 $\pm$ 0.004&3&0.103 $\pm$ 0.003&8&0.149 $\pm$ 0.008&4&0.143 $\pm$ 0.005&5\\
  			rcv1 (subset 1)&\textBF{0.462 $\pm$ 0.005}&\textBF{1}&0.443 $\pm$ 0.009&2&0.349 $\pm$ 0.010&7&0.351 $\pm$ 0.010&6&0.384 $\pm$ 0.009&5&0.344 $\pm$ 0.008&8&0.403 $\pm$ 0.007&3&0.394 $\pm$ 0.009&4\\
  			rcv1 (subset 2)&\textBF{0.461 $\pm$ 0.006}&\textBF{1}&0.425 $\pm$ 0.012&2&0.373 $\pm$ 0.016&6.5&0.373 $\pm$ 0.015&6.5&0.403 $\pm$ 0.012&5&0.362 $\pm$ 0.015&8&0.417 $\pm$ 0.010&3&0.411 $\pm$ 0.009&4\\
  			rcv1 (subset 3)&\textBF{0.461 $\pm$ 0.009}&\textBF{1}&0.387 $\pm$ 0.003&5&0.366 $\pm$ 0.009&6&0.365 $\pm$ 0.010&7&0.396 $\pm$ 0.012&4&0.359 $\pm$ 0.010&8&0.423 $\pm$ 0.011&2&0.409 $\pm$ 0.008&3\\
  			bibtex&0.419 $\pm$ 0.015&5&\textBF{0.474 $\pm$ 0.004}&\textBF{1}&0.390 $\pm$ 0.006&8&0.396 $\pm$ 0.007&6.5&0.420 $\pm$ 0.004&4&0.396 $\pm$ 0.007&6.5&0.421 $\pm$ 0.005&3&0.424 $\pm$ 0.002&2\\
  			corel16k001&0.256 $\pm$ 0.007&2&\textBF{0.274 $\pm$ 0.007}&\textBF{1}&0.071 $\pm$ 0.003&4.5&0.070 $\pm$ 0.003&6&0.137 $\pm$ 0.004&3&0.071 $\pm$ 0.004&4.5&0.052 $\pm$ 0.007&7&0.049 $\pm$ 0.003&8\\
  			delicious&0.230 $\pm$ 0.003&3&\textBF{0.304 $\pm$ 0.002}&\textBF{1}&0.194 $\pm$ 0.003&6.5&0.194 $\pm$ 0.003&6.5&0.241 $\pm$ 0.002&2&0.172 $\pm$ 0.002&8&0.211 $\pm$ 0.013&5&0.226 $\pm$ 0.003&4\\
  			mediamill&\textBF{0.581 $\pm$ 0.003}&\textBF{1}&0.545 $\pm$ 0.002&5&0.547 $\pm$ 0.002&3.5&0.547 $\pm$ 0.002&3.5&0.562 $\pm$ 0.002&2&0.543 $\pm$ 0.002&6&0.498 $\pm$ 0.023&8&0.520 $\pm$ 0.003&7\\
  			bookmarks&0.257 $\pm$ 0.044&&\textBF{0.281 $\pm$ 0.004}&&0.201$\pm$0.002&&\textBF{$-$}&&\textBF{$-$}&&0.202 $\pm$ 0.002&&0.1801 $\pm$ 0.003&&\textbf{$-$}&\\
  			\hline
  		\end{tabular}
  	}
  	\label{Err:AccuExamFMacroFMicroF}
  \end{table*}    
  The regularization parameters $\alpha$ and $\beta$ are tuned in the range given previously. The plots of Figure~\ref{groupVsNoGroup} shows the classification performance of GroPLE in terms of \textit{accuracy} and \textit{micro $f_1$}.  It can be seen from  Figure~\ref{groupVsNoGroup} that the classification performance of GroPLE is nearly constant for different group size when the latent dimension $d$ is small which is also obvious as there are less number of features to differentiate between one group from others. The classification performance is improved as we increase the number of groups $K$ for sufficiently large $d$.   It can also be seen from Figure~\ref{fig:groupVsNoGroupMicroF1} the performance degrade for sufficiently large group size $K$. Hence, by considering the balance between latent dimension $d$ and number of groups $K$, we fixed the value of $K$ and $d$ to $10$ and $100$, respectively, for subsequent experiments.

Table~\ref{Err:AccuExamFMacroFMicroF} gives the comparative analysis of the proposed method GroPLE against state-of-the-art algorithms on  twelve data sets. We have conducted \textit{five-fold cross validation} and the $mean$, $std$ and $rank$ is recorded.  For any data set and given evaluation metric where two or more algorithms obtain the same performance, the rank of these algorithm are assigned with the average rank of them. Furthermore, the best results among all the algorithms being compared are highlighted in boldface. For each data set, the number of latent dimension space $d$ is fixed to $100$ and the number of groups $K$ is set to $10$.  The regularization parameter $\lambda_1$ and $\lambda_2$ are fixed to $0.001$ and $1$, respectively, and the parameters $\alpha$ and $\beta$ are searched in the range given previously.

\begin{table}[ht]
	\renewcommand{\arraystretch}{1.2}
	\centering
	\caption{Summary of the Friedman Statistics $F_F(\mathcal{K}=8,\mathcal{N}=11)$ and the	Critical Value in Terms of Each Evaluation Metric($\mathcal{K}$: \# Comparing Algorithms; $\mathcal{N}$: \# Data Sets).}
	\begin{tabular}{llc}
		\toprule
		Metric &$F_F$&Critical Value ($\alpha = 0.05$)\\
		\toprule
		Accuracy&7.758		
		&\multirow{4}{*}{2.143}\\
		Example Base $F_1$&8.810&\\
		Macro $F_1$&7.234&\\
		Micro $F_1$&4.318&\\
		\hline
	\end{tabular}
	\label{ffTest}
\end{table}
 To conduct statistical performance analysis among the algorithms being compared, we employed \textit{Friedman test}\footnote{\label{excludeBookmarks} Results of bookmarks data set is not included for this test.} 
 which is a favorable statistical test for comparing more than two algorithms over multiple data sets~\cite{demvsar2006statistical}. 
 \begin{figure}[ht!]
 	\centering
 	\begin{subfigure}{.5\textwidth}
 		\centering
 		\includegraphics[width=3in,height=2.1in]{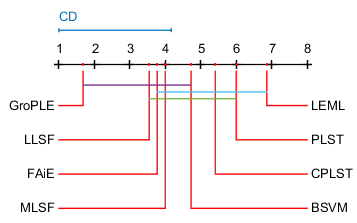}
 		\caption{Accuracy}
 		\label{fig:sub2}
 	\end{subfigure}%
 	\begin{subfigure}{.5\textwidth}
 		\centering
 		\includegraphics[width=3in,height=2.1in]{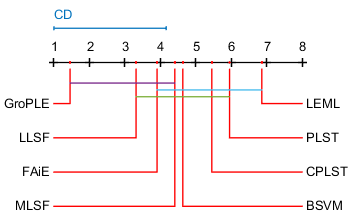}
 		\caption{Example based F$\textsubscript{1}$ }
 		\label{fig:sub4}
 	\end{subfigure}\\
 	\begin{subfigure}{.5\textwidth}
 		\centering
 		\includegraphics[width=3in,height=2.1in]{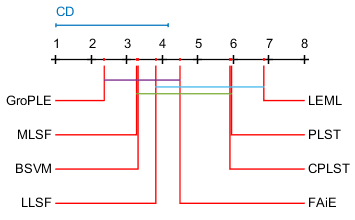}
 		\caption{Macro F$\textsubscript{1}$ }
 		\label{fig:sub5}%
 	\end{subfigure}%
 	\begin{subfigure}{.5\textwidth}
 		\centering
 		\includegraphics[width=3in,height=2.1in]{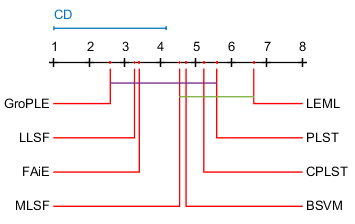}
 		\caption{Micro F$\textsubscript{1}$ }
 		\label{fig:sub6}
 	\end{subfigure}
 	\caption{CD diagrams of the comparing algorithms under each evaluation criterion.}
 	\label{CDDiagram}
 \end{figure} 
 Table \ref{ffTest} provides the \textit{Friedman statistics} $F_F$ and the corresponding critical value in terms of each evaluation metric.
 As shown in Table \ref{ffTest} at significance level $\alpha = 0.05$, Friedman test rejects the null hypothesis of “equal” performance for each evaluation metric. This leads to the use of post-hoc tests for pairwise comparisons. The Nemenyi test~\cite{demvsar2006statistical} is employed to test whether our proposed method GroPLE achieves a competitive performance against the algorithms  being compared.  The performance of two classifiers is significantly different if the corresponding average ranks differ by at least the critical difference $CD = q_{\alpha}\sqrt{ \frac{\mathcal{K}(\mathcal{K}+1)}{6\mathcal{N}}}$. At significance level $\alpha = 0.05$, the value of $q_{\alpha}= 3.031$, for Nemenyi test with $\mathcal{K}=8$~\cite{demvsar2006statistical}, and thus $CD=3.1658$. Figure~\ref{CDDiagram} gives the CD diagrams~\cite{demvsar2006statistical} for each evaluation criterion, where the average rank of each comparing algorithm is marked along the axis (lower ranks to the left). It can be seen from the Figure~\ref{CDDiagram} that the proposed method GroPLE achieve  better performance as compared to the other algorithms in terms of each evaluation metric.

Furthermore, we have also evaluated the approximation of label matrix $Y$ by comparing different label embedding based comparing algorithms. On each data set, we have conducted \textit{five-fold cross validation} and the $mean$, $std$ and $rank$ is recorded. Table~\ref{Err:LSDRapprox} reports the approximation result in terms of different evaluation metrics. It can be seen from the Table~\ref{Err:LSDRapprox} that the proposed method GroPLE is ranked higher among the algorithms being compared. It can also be seen that the approximation result of PLST is better than FAiE but the same is not being reflected in Table~\ref{Err:AccuExamFMacroFMicroF}. In FAiE, the goal is towards learning a predictable representation of label vectors by using the correlation that exists in the feature space. This prompts a new line of future research in which we plan to incorporate feature space correlation in GroPLE while learning the label embedding.
\begin{table*}[ht!]
	\renewcommand{\arraystretch}{1.2}
	\centering	
	\captionsetup{font=scriptsize,justification=centering}
	\caption{Approximation degree of each label embedding based comparing algorithm (mean$\pm$std rank)  in terms of Accuracy, Example Based $F_1$, Macro $F_1$, and Micro $F_1$. Method that cannot be run  with available resources are denoted as ``-".}
	\adjustbox{max width=\linewidth}{
		\begin{tabular}{lclclclcl||clclclcl}
			\hline
			\multirow{2}{*}{Data set}&\multicolumn{8}{c||}{Accuracy}&\multicolumn{8}{c}{Example based $F_1$} \\ \cline{2-17}
			&GroPLE&&PLST&&CPLST&&FAiE&&~~GroPLE&&PLST&&CPLST&&FAiE& \\
			\hline
			genbase&\textBF{1.000 $\pm$ 0.000}&\textBF{1}&0.995 $\pm$ 0.001&2.5&0.995 $\pm$ 0.001&2.5&0.994 $\pm$ 0.001&4&~~\textBF{1.000 $\pm$ 0.000}&\textBF{1}&0.997 $\pm$ 0.001&2&0.996 $\pm$ 0.000&3.5&0.996 $\pm$ 0.001&3.5\\
			medical&\textBF{1.000 $\pm$ 0.000}&\textBF{1}&0.989 $\pm$ 0.001&4&0.993 $\pm$ 0.000&2.5&0.993 $\pm$ 0.000&2.5&~~\textBF{1.000 $\pm$ 0.000}&\textBF{1}&0.992 $\pm$ 0.001&4&0.994 $\pm$ 0.000&2.5&0.994 $\pm$ 0.000&2.5\\
			CAL500&0.984 $\pm$ 0.001&2&\textBF{1.000 $\pm$ 0.000}&\textBF{1}&0.755 $\pm$ 0.003&3&0.548 $\pm$ 0.003&4&~~0.992 $\pm$ 0.000&2&\textBF{1.000 $\pm$ 0.000}&\textBF{1}&0.856 $\pm$ 0.002&3&0.700 $\pm$ 0.003&4\\
			corel5k&0.847 $\pm$ 0.001&3&\textBF{0.947 $\pm$ 0.001}&1&0.917 $\pm$ 0.003&2&0.530 $\pm$ 0.002&4&~~0.901 $\pm$ 0.001&3&\textBF{0.967 $\pm$ 0.001}&\textBF{1}&0.948 $\pm$ 0.002&2&0.644 $\pm$ 0.002&4\\
			rcv1 (subset 1)&\textBF{1.000 $\pm$ 0.000}&\textBF{1}&0.986 $\pm$ 0.000&2&0.980 $\pm$ 0.001&3&0.903 $\pm$ 0.001&4&~~\textBF{1.000 $\pm$ 0.000}&\textBF{1}&0.992 $\pm$ 0.000&2&0.988 $\pm$ 0.001&3&0.937 $\pm$ 0.000&4\\
			rcv1 (subset 2)&\textBF{1.000 $\pm$ 0.000}&\textBF{1}&0.989 $\pm$ 0.000&2.5&0.989 $\pm$ 0.001&2.5&0.902 $\pm$ 0.002&4&~~\textBF{1.000 $\pm$ 0.000}&\textBF{1}&0.994 $\pm$ 0.000&2.5&0.994 $\pm$ 0.000&2.5&0.933 $\pm$ 0.001&4\\
			rcv1 (subset 3)&\textBF{1.000 $\pm$ 0.000}&\textBF{1}&0.996 $\pm$ 0.000&2&0.964 $\pm$ 0.002&3&0.894 $\pm$ 0.001&4&~~\textBF{1.000 $\pm$ 0.000}&\textBF{1}&0.998 $\pm$ 0.000&2&0.977 $\pm$ 0.001&3&0.926 $\pm$ 0.001&4\\
			bibtex&0.871 $\pm$ 0.004&3&0.974 $\pm$ 0.003&2&\textBF{0.975 $\pm$ 0.002}&\textBF{1}&0.764 $\pm$ 0.002&4&~~0.897 $\pm$ 0.004&3&\textBF{0.980 $\pm$ 0.002}&1.5&\textBF{0.980 $\pm$ 0.002}&1.5&0.801 $\pm$ 0.002&4\\
			corel16k001&0.953 $\pm$ 0.001&2&\textBF{0.983 $\pm$ 0.000}&\textBF{1}&0.908 $\pm$ 0.002&3&0.202 $\pm$ 0.003&4&~~0.969 $\pm$ 0.001&2&\textBF{0.990 $\pm$ 0.000}&\textBF{1}&0.937 $\pm$ 0.001&3&0.271 $\pm$ 0.004&4\\
			delicious&0.649 $\pm$ 0.001&3&\textBF{0.992 $\pm$ 0.000}&\textBF{1}&0.836 $\pm$ 0.001&2&0.372 $\pm$ 0.001&4&~~0.774 $\pm$ 0.001&3&\textBF{0.996 $\pm$ 0.000}&\textBF{1}&0.906 $\pm$ 0.000&2&0.513 $\pm$ 0.001&4\\
			mediamill&\textBF{1.000 $\pm$ 0.000}&\textBF{1}&0.912 $\pm$ 0.001&3&0.927 $\pm$ 0.001&2&0.804 $\pm$ 0.001&4&~~\textBF{1.000 $\pm$ 0.000}&\textBF{1}&0.931 $\pm$ 0.001&3&0.940 $\pm$ 0.001&2&0.860 $\pm$ 0.001&4\\
			\hline
			Average Rank&&1.73&&2&&2.41&&3.86&&1.73&&1.91&&2.55&&3.82\\ 
			Total Order&GroPLE&$\succ$&PLST&$\succ$&CPLST&$\succ$&FAiE&&GroPLE&$\succ$&PLST&$\succ$&CPLST&$\succ$&FAiE\\
			\hline  
			\multirow{2}{*}{Data set}&\multicolumn{8}{c||}{Macro $F_1$}&\multicolumn{8}{c}{Micro $F_1$} \\ \cline{2-17}
			&GroPLE&&PLST&&CPLST&&FAiE&&GroPLE&&PLST&&CPLST&&FAiE& \\
			\hline
			genbase&\textBF{1.000 $\pm$ 0.000}&\textBF{1}&0.879 $\pm$ 0.017&3&0.881 $\pm$ 0.020&2&0.853 $\pm$ 0.025&4&~~\textBF{1.000 $\pm$ 0.000}&\textBF{1}&0.997 $\pm$ 0.001&2&0.996 $\pm$ 0.001&3&0.995 $\pm$ 0.001&4\\
			medical&\textBF{1.000 $\pm$ 0.000}&\textBF{1}&0.716 $\pm$ 0.012&4&0.809 $\pm$ 0.006&2&0.805 $\pm$ 0.011&3&~~\textBF{1.000 $\pm$ 0.000}&\textBF{1}&0.992 $\pm$ 0.001&4&0.995 $\pm$ 0.000&2.5&0.995 $\pm$ 0.000&2.5\\
			CAL500&0.893 $\pm$ 0.004&2&\textBF{1.000 $\pm$ 0.003}&\textBF{1}&0.502 $\pm$ 0.003&3&0.332 $\pm$ 0.004&4&~~0.992 $\pm$ 0.000&2&\textBF{1.000 $\pm$ 0.000}&\textBF{1}&0.861 $\pm$ 0.002&3&0.713 $\pm$ 0.002&4\\
			corel5k&0.337 $\pm$ 0.003&3&\textBF{0.569 $\pm$ 0.003}&\textBF{1}&0.482 $\pm$ 0.006&2&0.112 $\pm$ 0.001&4&~~0.918 $\pm$ 0.001&3&\textBF{0.973 $\pm$ 0.001}&\textBF{1}&0.958 $\pm$ 0.002&2&0.695 $\pm$ 0.002&4\\
			rcv1 (subset 1)&\textBF{1.000 $\pm$ 0.000}&\textBF{1}&0.772 $\pm$ 0.005&2&0.730 $\pm$ 0.006&3&0.606 $\pm$ 0.004&4&~~\textBF{1.000 $\pm$ 0.000}&\textBF{1}&0.991 $\pm$ 0.000&2&0.986 $\pm$ 0.001&3&0.940 $\pm$ 0.001&4\\
			rcv1 (subset 2)&\textBF{1.000 $\pm$ 0.000}&\textBF{1}&0.817 $\pm$ 0.008&3&0.820 $\pm$ 0.006&2&0.618 $\pm$ 0.006&4&~~\textBF{1.000 $\pm$ 0.000}&\textBF{1}&0.992 $\pm$ 0.000&2.5&0.992 $\pm$ 0.001&2.5&0.938 $\pm$ 0.001&4\\
			rcv1 (subset 3)&\textBF{1.000 $\pm$ 0.000}&\textBF{1}&0.893 $\pm$ 0.008&2&0.668 $\pm$ 0.006&3&0.598 $\pm$ 0.004&4&~~\textBF{1.000 $\pm$ 0.000}&\textBF{1}&0.997 $\pm$ 0.000&2&0.973 $\pm$ 0.001&3&0.934 $\pm$ 0.001&4\\
			bibtex&0.821 $\pm$ 0.006&3&0.970 $\pm$ 0.004&2&\textBF{0.973 $\pm$ 0.004}&\textBF{1}&0.761 $\pm$ 0.004&4&~~0.937 $\pm$ 0.002&3&\textBF{0.988 $\pm$ 0.001}&\textBF{1.5}&\textBF{0.988 $\pm$ 0.001}&\textBF{1.5}&0.871 $\pm$ 0.002&4\\
			corel16k001&0.753 $\pm$ 0.004&2&\textBF{0.883 $\pm$ 0.002}&\textBF{1}&0.636 $\pm$ 0.006&3&0.044 $\pm$ 0.001&4&~~0.975 $\pm$ 0.001&2&\textBF{0.991 $\pm$ 0.000}&\textBF{1}&0.952 $\pm$ 0.001&3&0.325 $\pm$ 0.004&4\\
			delicious&0.223 $\pm$ 0.001&3&\textBF{0.969 $\pm$ 0.002}&\textBF{1}&0.476 $\pm$ 0.002&2&0.122 $\pm$ 0.001&4&~~0.802 $\pm$ 0.001&3&\textBF{0.997 $\pm$ 0.000}&\textBF{1}&0.913 $\pm$ 0.000&2&0.553 $\pm$ 0.001&4\\
			mediamill&\textBF{1.000 $\pm$ 0.000}&\textBF{1}&0.379 $\pm$ 0.003&3&0.505 $\pm$ 0.004&2&0.191 $\pm$ 0.001&4&~~\textBF{1.000 $\pm$ 0.000}&\textBF{1}&0.969 $\pm$ 0.000&3&0.980 $\pm$ 0.000&2&0.898 $\pm$ 0.001&4\\
			\hline
			Average Rank&&1.73&&2.09&&2.27&&3.91&&1.73&&1.91&&2.5&&3.86\\
			Total Order&GroPLE&$\succ$&PLST&$\succ$&CPLST&$\succ$&FAiE&&GroPLE&$\succ$&PLST&$\succ$&CPLST&$\succ$&FAiE\\
			\hline
		\end{tabular}
	}
	\label{Err:LSDRapprox}
\end{table*}

 \begin{figure}[ht!]
	\centering
	\adjustbox{max width=\linewidth}{
		\begin{subfigure}{.5\textwidth}
			\centering
			\includegraphics[width=3.25in,height=2.5in]{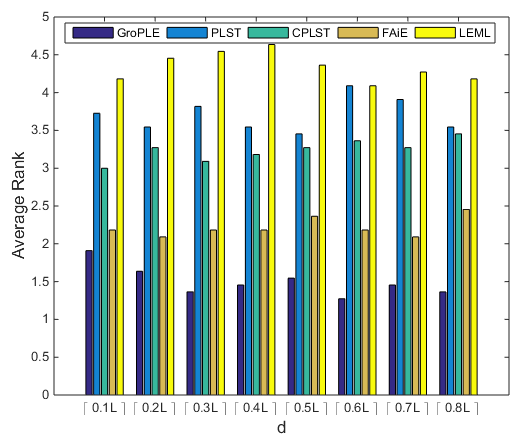}
			\caption{Accuracy}
			\label{fig:avgRankAccuracy}
		\end{subfigure}%
		\begin{subfigure}{.5\textwidth}
			\centering
			\includegraphics[width=3.2in,height=2.5in]{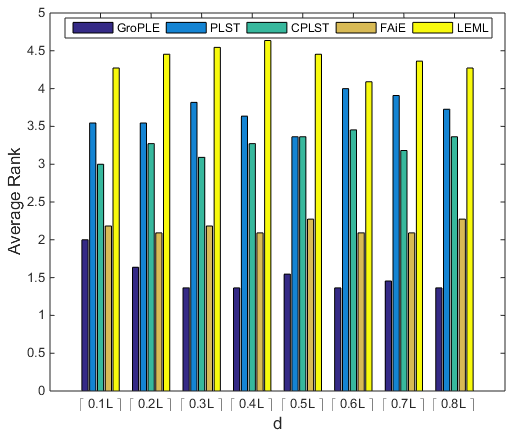}
			\caption{Example based F$\textsubscript{1}$ }
			\label{fig:avgRankExampleBasedF1}
	\end{subfigure}}\\
	\adjustbox{max width=\linewidth}{
		\begin{subfigure}{.5\textwidth}
			\centering
			\includegraphics[width=3.2in,height=2.5in]{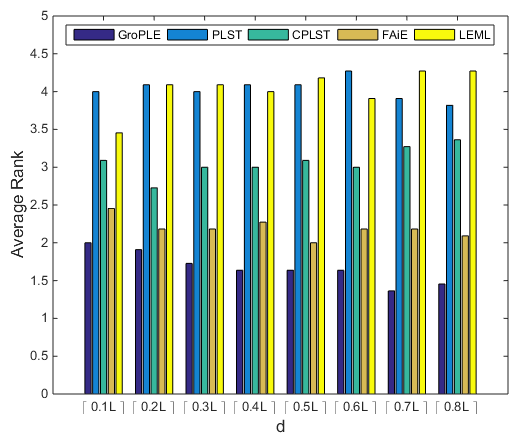}
			\caption{Macro F$\textsubscript{1}$ }
			\label{fig:avgRankMacroF1}%
		\end{subfigure}%
		\begin{subfigure}{.5\textwidth}
			\centering
			\includegraphics[width=3.2in,height=2.5in]{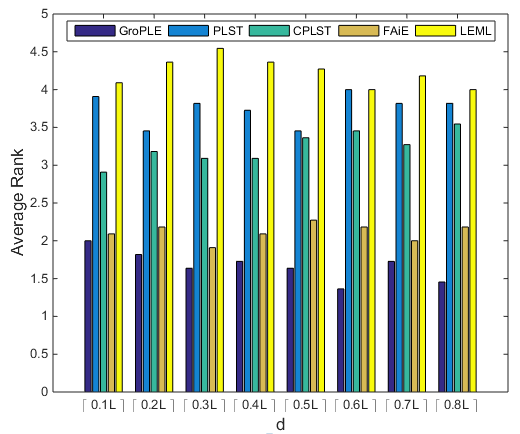}
			\caption{Micro F$\textsubscript{1}$ }
			\label{fig:avgRankMicroF1}
		\end{subfigure}
	}
	\caption{Average rank of each embedding based comparing algorithm for different values of $d$.}
	\label{avgRank}
\end{figure}
We have also demonstrated the effect of the reduced dimension space $d$ that varied from $\{\lceil 0.1L \rceil, \lceil 0.2L \rceil, \dots, \lceil 0.8L \rceil\}$ for each embedding based comparing algorithm. For each value of $d$, the comparing algorithm's parameters are searched in the range given previously. On each data set, we have conducted \textit{five-fold cross validation} for each value of $d$ and the \textit{rank} is recorded. For any (data set, $d$)-pair and given evaluation metric, the algorithms are ranked in a similar way as shown in Table~\ref{Err:AccuExamFMacroFMicroF}. We report the average rank of each comparing algorithms on eleven data sets
\textsuperscript{\ref{excludeBookmarks}}
in Figure~\ref{avgRank}. It can be seen that for each value of $d$, the proposed method GroPLE is ranked better than that of other algorithms in terms of each evaluation metric.

\begin{figure}[ht!]
		\centering
		\includegraphics[width=4.5in,height=4in]{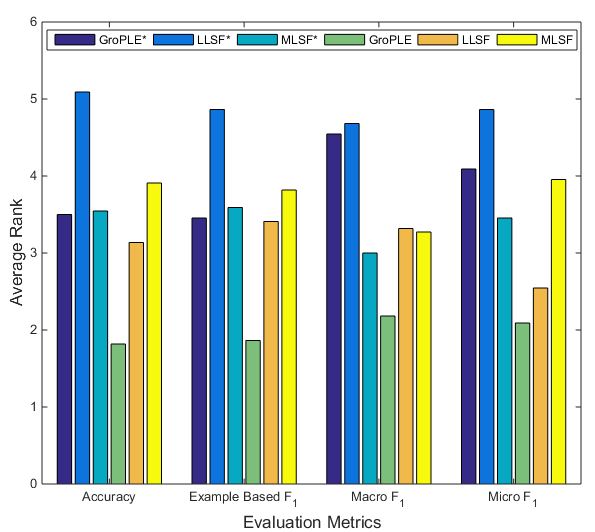}
		\caption{Average rank of GroPLE, LLSF and MLSF for two different combinations of $\alpha$ and $\beta$.}
		\label{GroPLE_LLSF_MLSF}		
\end{figure}

The proposed method GroPLE uses LSSF to map the original data point to their low-dimensional representation obtained as a result of label embedding process. In LLSF, the mapping is achieved by minimizing a loss function (Eq.~\ref{featureEmbedding}) which consists of components to measure the approximation error and to preserve the correlation information that exists in the embedded space (or, label space). A similar formulation for feature space embedding is used in MLSF to learn the meta-label specific features. The difference lies precisely in the use of components to  preserve the correlation in the embedded space. In MLSF,  the authors propose the use of group information present in the original feature and label space to learn the meta-labels. Once the meta-labels are known, a feature-space embedding is learnt from the feature space to the meta-label space without guaranteeing preservation of correlation between the meta-labels. To observe the effect of correlation component, we modify the objective function of MLSF and include a component to preserve the correlation between the meta-labels. For fair comparison with GroPLE and LLSF, we further adopt Accelerated Proximal Gradient search~\cite{toh2010accelerated} to optimize the objective function for feature space embedding. We have considered two different combinations - one without the component to preserve correlation i.e., with $\alpha = 0$ and the other with $\alpha > 0$. For both the combinations, the parameter $\beta$ is tuned in the range given previously. On each data set, we have conducted \textit{five-fold cross validation} for each combination of $\alpha$ and $\beta$ i.e., ($\alpha = 0$, $\beta > 0$) and ($\alpha > 0$, $\beta > 0$) and recorded the \textit{rank}. Figure~\ref{GroPLE_LLSF_MLSF} depicts the average rank obtained by the comparing algorithms. For simplicity of representation, we have used $A^*$ to denote that the feature space embedding is performed without the correlation components in method $A$. It can be seen from  Figure~\ref{GroPLE_LLSF_MLSF} that the performance of GroPLE and LLSF have improved when the embedding process is guided by the correlation information. It can also be seen that the performance of MLSF has slightly decreased after the meta-labels correlation is incorporated. This is due to the reason that the meta-labels are formed by converting the label groups to equivalent decimal value. This conversion helps capture the statistics of a group (meta-label) so that the group specific feature can be learnt without capturing the statistics of other groups.

\section{Conclusions}
	\label{conclusion}
	This paper presented a new multi-label classification method, called GroPLE, which embeds the original label vectors to  a low-dimensional space by retaining the group dependencies. We ensure that labels belonging to the same group share the same sparsity pattern in their low-dimensional representations. In order to achieve the embedding of feature vectors, a linear mapping is then learnt that maps the feature vectors onto the same set of points which are obtained as a result of label embedding phase. We achieve this by a separate optimization problem.  Extensive comparative studies validate the effectiveness of GroPLE against the state-of-the-art multi-label learning approaches.
	
	In the future, it is interesting to see whether GroPLE can be further improved by considering side information from feature space while label embedding. Furthermore, designing other ways to fulfill the strategy of group formation and modeling group-specific label embedding is a direction worth studying.

\section*{Acknowledgements}
We thank the anonymous reviewers whose comments/suggestions helped improve and clarify this manuscript to a large extent. 

\newpage	
	\appendix
		\section{Derivation of the Criterion in Eq. (\ref{GroPLE})}
		\label{LSDRderivation}
		For simplicity of notation, the matrix formed by arranging the columns of $V^k$, $1\le k \le K$, according to indices of columns in $Y$ will be referred to as $V$ in subsequent discussion, $f(U, V)$ is written as $f_V(U)$ when $V$ is held constant and  $f_U(V)$ when $U$ is held constant. For given $V$, the factor matrix $U$ can be obtained by solving the following subproblem	
		\begin{equation}
		\label{eqUpdateU}
		\underset{U}{min}~f_V(U)=~\|Y-UV\|^2_F + \lambda_1\|U\|_F^2 + c
		\end{equation}
		where $c \ge 0$ is a constant. The subproblem given in Eq.~(\ref{eqUpdateU}) has a closed form solution. Taking the derivative of $f_V(U)$ w.r.t $U$, and setting the derivative (in matrix notation) to zero, we have
		\begin{align}
		\nabla f_V(U) &= 2((Y-UV)(-V^T) + \lambda_1 U)= 0 \nonumber \\
		&\Rightarrow U = YV^T \inv{(VV^T+\lambda_1 I)}
		\end{align}
		For fixed $U$, the matrix $V^k$, $k \in \{1, \dots, K\}$, can be obtained by solving the following subproblem
		\begin{equation}
		\label{subproblemVg}
		\underset{V^k}{min}~f_{U}(V^k) = \|Y^{k}-UV^{k}\|^2_F  + \lambda_2\|V^k\|_{2,1} + c
		\end{equation}
		The above objective function is a composite convex function involving the sum of a smooth and a non-smooth function  of the form
		\begin{equation}
		\label{compositeFun}
		\underset{V^k}{min}~f_{U}(V^k) = g(V^k) + h(V^k)
		\end{equation}
		where $g(V^k) = \|Y^{k}-UV^{k}\|^2_F $ is convex and differentiable and $h(V^k) = \lambda_2\|V^k\|_{2,1}$ is closed, convex but non-differentiable. 
		
		We further show that for any two matrices ${V^{'}}^{k}$, ${V^{''}}^{k} \in \mathbb{R}^{d \times L_k}$, the function $g(V^k) $ is Lipschitz continuous. 
		The gradient of $g(V^k)$ (in matrix notation) is given by
		\begin{align*}
		\nabla g(V^k) = 2(U^{T}UV^k - U^{T}Y)
		\end{align*}
		For any two matrices ${V^{'}}^{k}$ and ${V^{''}}^{k}$, we have
		\begin{align*}
		\|\nabla g({V^{'}}^{k}) - \nabla g({V^{''}}^{k}) \|_F^2 &= \|2(U^{T}U{V^{'}}^{k} - U^{T}Y) - 2(U^{T}U{V^{''}}^{k} - U^{T}Y)\|^2_F \\
		&=\|2U^{T}U({V^{'}}^{k}-{V^{''}}^{k})\|_F^2\\
		&\le\|2U^{T}U\|^2_F~\|{V^{'}}^{k} - {V^{''}}^{k}\|_F^2
		\end{align*}	
		Therefore, the Lipschitz constant is
		\begin{equation}
		L_g = \sqrt{\|2U^{T}U\|^2_F} 
		\end{equation}	
		
		We employ Accelerated Proximal Gradient search~\cite{toh2010accelerated} which is specifically tailored to minimize the optimization problem given in Eq.~(\ref{compositeFun}). Such an optimization strategy is suitable in the present situation as the computation of the proximal operation is inexpensive. The optimization step of Accelerated Proximal Gradient iterates as follows
		\begin{align}
		G_t &= {V^k}^{(t)} + \frac{b_{t-1}-1}{b_t}({V^k}^{(t)} - {V^k}^{(t-1)})\\
		{V^k}^{(t)} &= prox_{h}(G_t - \frac{1}{L_g}\nabla g(G_t)) 
		\end{align}
		
		It is shown in~\cite{toh2010accelerated} that setting $b_t$ satisfying $b_t^2 - b_t \le b_{t-1}^2$ can improve convergence rate to $O(\frac{1}{t^2})$, ${V^k}^{(t)}$ is the result of $t$th iteration. Proximal mapping of a convex function $h$ is given by
		\begin{equation}
		prox_h(V^k) = \underset{W}{argmin}~\left(h(W) + \frac{1}{2}\|W-V^k\|^2_2~\right)
		\end{equation} 	
		
		In the present situation, where $h(V^k) = \lambda_2\|V^k\|_{2,1}$, $prox_h(V^k)$ is the shrinkage function $S[\cdot]$ and is given by
		\begin{equation}
		S_{\frac{\lambda_2}{L_g}}[V^k] = \bigg [\frac{v^{k}_{i}}{\|v^{k}_{i}\|_2} (\|v^{k}_{i}\|_2- \lambda_2/L_g)_{+}\bigg ]_{i=1}^{i=d}
		\end{equation}
		where $(z)_{+} = max(z,0)$ and $v^k_i$ is the $i$th row of $V^k$. Algorithm~\ref{algo:LE} outlines the main flow of the optimization steps to solve Eq.~(\ref{GroPLE}).   
		\begin{algorithm}[ht!]
			\SetAlgoLined
			\SetKwData{Left}{left}\SetKwData{This}{this}\SetKwData{Up}{up}
			\SetKwFunction{Union}{Union}\SetKwFunction{FindCompress}{FindCompress}
			\SetKwInOut{Input}{input}\SetKwInOut{Output}{output}
			\Input{Label Matrix: $Y$, Size of Latent Dimension Space: $d$, Number of Groups: $K$, Regularization Parameters: $\lambda_1$ and $\lambda_2$}
			\Output{Basis Matrix: $U$, Coefficient Matrix: $V$}
			\BlankLine
			\textbf{initialize :} $U$\\
			Form label groups $Y^1$, $Y^2$, \dots, $Y^K$ \\
			\Repeat{stop criterion reached}{
				\For{k $\in$ \{1,\dots, K\}}{
					$V^k \leftarrow$ APG($U$, $Y^k$, $\lambda_2$)
				}
				$V \leftarrow Combine(V^1, V^2, \dots, V^k)$ \\
				$U \leftarrow YV^T \inv{(VV^T+\lambda_1 I)}$ 
			}
			\caption{Label-Embedding ( $Y$, $d$, $K$, $\lambda_1$, $\lambda_2$)}
			\label{algo:LE}
		\end{algorithm}    
		
		\begin{algorithm}[ht!]
			\SetAlgoLined
			\SetKwData{Left}{left}\SetKwData{This}{this}\SetKwData{Up}{up}
			\SetKwFunction{Union}{Union}\SetKwFunction{FindCompress}{FindCompress}
			\SetKwInOut{Input}{input}\SetKwInOut{Output}{output}
			\Input{Basic Matrix: $U$, Label Matrix: $Y^k$ and Regularization Parameters: $\lambda_2$}
			\Output{Coefficient Matrix: $V^k$}
			\BlankLine
			\textbf{initialize :} \\
			$b_0$, $b_1 \leftarrow 1$, $V^k_0$, $V^k_1 \leftarrow$ \inv{($U^TU + \gamma I $)}$U^TY^k$\\
			\Repeat{stop criterion reached}{
				$G_t \leftarrow {V^k}^{(t)} + \frac{b_{t-1}-1}{b_t}({V^k}^{(t)} - {V^k}^{(t-1)})$ \\
				${V^k}^{(t)} = S_{\frac{\lambda_2}{L_g}}[G_t - \frac{1}{L_g}\nabla g(G_t)]$\\
				$b_t \leftarrow \frac{1+ \sqrt{1 + 4b_t^2}}{2}$ \\
				$t \leftarrow t+1$
			}
			$V^k \leftarrow V^k_t$
			\caption{APG ($U$, $Y^k$, $\lambda_2$)}
			\label{Algo:APG}
		\end{algorithm} 
	
	\newpage
	\bibliographystyle{plain}
	\bibliography{ijcai17references}

\begin{thebibliography}{10}

\bibitem{balasubramanian2012landmark}
Krishnakumar Balasubramanian and Guy Lebanon.
\newblock The landmark selection method for multiple output prediction.
\newblock {\em arXiv preprint arXiv:1206.6479}, 2012.

\bibitem{beck2009fast}
Amir Beck and Marc Teboulle.
\newblock A fast iterative shrinkage-thresholding algorithm for linear inverse
  problems.
\newblock {\em SIAM journal on imaging sciences}, 2(1):183--202, 2009.

\bibitem{bhatia2015sparse}
Kush Bhatia, Himanshu Jain, Purushottam Kar, Manik Varma, and Prateek Jain.
\newblock Sparse local embeddings for extreme multi-label classification.
\newblock In {\em Advances in Neural Information Processing Systems}, pages
  730--738, 2015.

\bibitem{bi2013efficient}
Wei Bi and James Kwok.
\newblock Efficient multi-label classification with many labels.
\newblock In {\em International Conference on Machine Learning}, pages
  405--413, 2013.

\bibitem{boutell2004learning}
Matthew~R Boutell, Jiebo Luo, Xipeng Shen, and Christopher~M Brown.
\newblock Learning multi-label scene classification.
\newblock {\em Pattern recognition}, 37(9):1757--1771, 2004.

\bibitem{cabral2011matrix}
Ricardo~Silveira Cabral, Fernando De~la Torre, Jo{\~a}o~Paulo Costeira, and
  Alexandre Bernardino.
\newblock Matrix completion for multi-label image classification.
\newblock In {\em NIPS}, volume 201, page~2, 2011.

\bibitem{CC01a}
Chih-Chung Chang and Chih-Jen Lin.
\newblock {LIBSVM}: A library for support vector machines.
\newblock {\em ACM Transactions on Intelligent Systems and Technology},
  2:27:1--27:27, 2011.

\bibitem{chen2012feature}
Yao-Nan Chen and Hsuan-Tien Lin.
\newblock Feature-aware label space dimension reduction for multi-label
  classification.
\newblock In {\em Advances in Neural Information Processing Systems}, pages
  1529--1537, 2012.

\bibitem{clareknowledge}
Amanda Clare and Ross~D King.
\newblock Knowledge discovery in multi-label phenotype data.
\newblock In {\em PKDD}, pages 42--53. Springer, 2001.

\bibitem{demvsar2006statistical}
Janez Dem{\v{s}}ar.
\newblock Statistical comparisons of classifiers over multiple data sets.
\newblock {\em JMLR}, 7(Jan):1--30, 2006.

\bibitem{fakhari2013combination}
Ali Fakhari and Amir Masoud~Eftekhari Moghadam.
\newblock Combination of classification and regression in decision tree for
  multi-labeling image annotation and retrieval.
\newblock {\em Applied Soft Computing}, 13(2):1292--1302, 2013.

\bibitem{furnkranz2008multilabel}
Johannes F{\"u}rnkranz, Eyke H{\"u}llermeier, Eneldo~Loza Menc{\'\i}a, and
  Klaus Brinker.
\newblock Multilabel classification via calibrated label ranking.
\newblock {\em Machine learning}, 73(2):133--153, 2008.

\bibitem{hsu2009multi}
Daniel~J Hsu, Sham Kakade, John Langford, and Tong Zhang.
\newblock Multi-label prediction via compressed sensing.
\newblock In {\em NIPS}, volume~22, pages 772--780, 2009.

\bibitem{huang2015learning}
Jun Huang, Guorong Li, Qingming Huang, and Xindong Wu.
\newblock Learning label specific features for multi-label classification.
\newblock In {\em Data Mining (ICDM), 2015 IEEE International Conference on},
  pages 181--190. IEEE, 2015.

\bibitem{huang2016learning}
Jun Huang, Guorong Li, Qingming Huang, and Xindong Wu.
\newblock Learning label-specific features and class-dependent labels for
  multi-label classification.
\newblock {\em IEEE Transactions on Knowledge and Data Engineering},
  28(12):3309--3323, 2016.

\bibitem{huang2012multi}
Sheng-Jun Huang, Zhi-Hua Zhou, and ZH~Zhou.
\newblock Multi-label learning by exploiting label correlations locally.
\newblock In {\em AAAI}, pages 949--955, 2012.

\bibitem{jain2010data}
Anil~K Jain.
\newblock Data clustering: 50 years beyond k-means.
\newblock {\em Pattern recognition letters}, 31(8):651--666, 2010.

\bibitem{jain1999data}
Anil~K Jain, M~Narasimha Murty, and Patrick~J Flynn.
\newblock Data clustering: a review.
\newblock {\em ACM computing surveys (CSUR)}, 31(3):264--323, 1999.

\bibitem{jian2016multi}
Ling Jian, Jundong Li, Kai Shu, and Huan Liu.
\newblock Multi-label informed feature selection.
\newblock In {\em 25th International Joint Conference on Artificial
  Intelligence}, pages 1627--1633, 2016.

\bibitem{jing2015semi}
Liping Jing, Liu Yang, Jian Yu, and Michael~K Ng.
\newblock Semi-supervised low-rank mapping learning for multi-label
  classification.
\newblock In {\em Proceedings of the IEEE Conference on Computer Vision and
  Pattern Recognition}, pages 1483--1491, 2015.

\bibitem{johnson1967hierarchical}
Stephen~C Johnson.
\newblock Hierarchical clustering schemes.
\newblock {\em Psychometrika}, 32(3):241--254, 1967.

\bibitem{karampatziakis2015scalable}
Nikos Karampatziakis and Paul Mineiro.
\newblock Scalable multilabel prediction via randomized methods.
\newblock {\em arXiv preprint arXiv:1502.02710}, 2015.

\bibitem{kimura2016simultaneous}
Keigo Kimura, Mineichi Kudo, and Lu~Sun.
\newblock Simultaneous nonlinear label-instance embedding for multi-label
  classification.
\newblock In {\em S+SSPR}, pages 15--25. Springer, 2016.

\bibitem{lin2014multi}
Zijia Lin, Guiguang Ding, Mingqing Hu, and Jianmin Wang.
\newblock Multi-label classification via feature-aware implicit label space
  encoding.
\newblock In {\em ICML}, pages 325--333, 2014.

\bibitem{ng2002spectral}
Andrew~Y Ng, Michael~I Jordan, and Yair Weiss.
\newblock On spectral clustering: Analysis and an algorithm.
\newblock In {\em Advances in neural information processing systems}, pages
  849--856, 2002.

\bibitem{prabhu2014fastxml}
Yashoteja Prabhu and Manik Varma.
\newblock Fastxml: A fast, accurate and stable tree-classifier for extreme
  multi-label learning.
\newblock In {\em Proceedings of the 20th ACM SIGKDD international conference
  on Knowledge discovery and data mining}, pages 263--272. ACM, 2014.

\bibitem{qian2010semi}
Buyue Qian and Ian Davidson.
\newblock Semi-supervised dimension reduction for multi-label classification.
\newblock In {\em AAAI}, volume~10, pages 569--574, 2010.

\bibitem{rai2015large}
Piyush Rai, Changwei Hu, Ricardo Henao, and Lawrence Carin.
\newblock Large-scale bayesian multi-label learning via topic-based label
  embeddings.
\newblock In {\em Advances in Neural Information Processing Systems}, pages
  3222--3230, 2015.

\bibitem{read2009classifier}
Jesse Read, Bernhard Pfahringer, Geoff Holmes, and Eibe Frank.
\newblock Classifier chains for multi-label classification.
\newblock In {\em ECML PKDD}, pages 254--269. Springer, 2009.

\bibitem{schapire2000boostexter}
Robert~E Schapire and Yoram Singer.
\newblock Boostexter: A boosting-based system for text categorization.
\newblock {\em Machine learning}, 39(2-3):135--168, 2000.

\bibitem{sorower2010literature}
Mohammad~S Sorower.
\newblock A literature survey on algorithms for multi-label learning.
\newblock {\em Oregon State University, Corvallis}, 2010.

\bibitem{spyromitros2008empirical}
Eleftherios Spyromitros, Grigorios Tsoumakas, and Ioannis Vlahavas.
\newblock An empirical study of lazy multilabel classification algorithms.
\newblock In {\em Hellenic conference on Artificial Intelligence}, pages
  401--406. Springer, 2008.

\bibitem{sun2016multi}
Lu~Sun, Mineichi Kudo, and Keigo Kimura.
\newblock Multi-label classification with meta-label-specific features.
\newblock In {\em Pattern Recognition (ICPR), 2016 23rd International
  Conference on}, pages 1612--1617. IEEE, 2016.

\bibitem{tai2012multilabel}
Farbound Tai and Hsuan-Tien Lin.
\newblock Multilabel classification with principal label space transformation.
\newblock {\em Neural Computation}, 24(9):2508--2542, 2012.

\bibitem{toh2010accelerated}
Kim-Chuan Toh and Sangwoon Yun.
\newblock An accelerated proximal gradient algorithm for nuclear norm
  regularized linear least squares problems.
\newblock {\em Pacific Journal of optimization}, 6(615-640):15, 2010.

\bibitem{tsoumakas2007random}
Grigorios Tsoumakas and Ioannis Vlahavas.
\newblock Random k-labelsets: An ensemble method for multilabel classification.
\newblock In {\em ECML}, pages 406--417. Springer, 2007.

\bibitem{von2007tutorial}
Ulrike Von~Luxburg.
\newblock A tutorial on spectral clustering.
\newblock {\em Statistics and computing}, 17(4):395--416, 2007.

\bibitem{wang2018learning}
Shangfei Wang, Shiyu Chen, Tanfang Chen, and Xiaoxiao Shi.
\newblock Learning with privileged information for multi-label classification.
\newblock {\em Pattern Recognition}, 81:60--70, 2018.

\bibitem{wang2014enhancing}
Shangfei Wang, Jun Wang, Zhaoyu Wang, and Qiang Ji.
\newblock Enhancing multi-label classification by modeling dependencies among
  labels.
\newblock {\em Pattern Recognition}, 47(10):3405--3413, 2014.

\bibitem{xu2016robust}
Chang Xu, Dacheng Tao, and Chao Xu.
\newblock Robust extreme multi-label learning.
\newblock In {\em Proceedings of the 22nd ACM SIGKDD International Conference
  on Knowledge Discovery and Data Mining, San Francisco, CA, USA August}, pages
  13--17, 2016.

\bibitem{yu2014large}
Hsiang-Fu Yu, Prateek Jain, Purushottam Kar, and Inderjit~S Dhillon.
\newblock Large-scale multi-label learning with missing labels.
\newblock In {\em ICML}, pages 593--601, 2014.

\bibitem{zelnik2004self}
Lihi Zelnik-Manor and Pietro Perona.
\newblock Self-tuning spectral clustering.
\newblock In {\em NIPS}, volume~17, page~16, 2004.

\bibitem{zhang2007ml}
Min-Ling Zhang and Zhi-Hua Zhou.
\newblock Ml-knn: A lazy learning approach to multi-label learning.
\newblock {\em Pattern recognition}, 40(7):2038--2048, 2007.

\bibitem{zhang2014review}
Min-Ling Zhang and Zhi-Hua Zhou.
\newblock A review on multi-label learning algorithms.
\newblock {\em IEEE transactions on knowledge and data engineering},
  26(8):1819--1837, 2014.

\end{thebibliography}

\end{document}